\documentclass[11pt]{article}
\usepackage[T1]{fontenc}
\usepackage[utf8]{inputenc}
\input{glyphtounicode}
\usepackage[final]{acl}

\usepackage{amsmath}
\usepackage{booktabs}
\usepackage{amssymb}

\usepackage{tikz}
\usetikzlibrary{shapes.geometric, arrows.meta, positioning, calc, shadows, fit, backgrounds}

\usepackage{multirow,booktabs}

\usepackage{times}
\usepackage{latexsym}

\usepackage{microtype}

\usepackage{inconsolata}

\usepackage{graphicx}
\usepackage{subcaption}

\usepackage[most]{tcolorbox}
\usepackage{xcolor}
\usepackage{ragged2e}
\usepackage{threeparttable}
\usepackage{hyperref}
\usepackage{algorithm}
\usepackage{algorithmic}

\usepackage{listings}

\lstdefinestyle{promptstyle}{
  basicstyle=\ttfamily\footnotesize,
  columns=fullflexible,
  breaklines=true,
  breakatwhitespace=true,
  keepspaces=true,
  showstringspaces=false
}

\newcommand{\drift}[1]{\textcolor{red!70!black}{\bfseries #1}}

\title{\textsc{SPASM}: Stable Persona-driven Agent Simulation for Multi-turn Dialogue Generation}

\author{%
  Han Luo $^{\spadesuit\blacksquare\blacklozenge}$\thanks{Han Luo worked on this study as a visiting student at the \href{https://labalab.li}{LabaLab (Language, Affect, and Behaviour in AI Lab)} in the Department of Industrial Engineering and Management at Ben-Gurion University of the Negev.} \quad
  Guy Laban $^{\blacklozenge}$\thanks{Corresponding author: \texttt{\href{mailto:laban@bgu.ac.il}{laban@bgu.ac.il}}}\\
  \\
  $^{\spadesuit}$ University of Leeds \quad
  $^{\blacksquare}$ Southwest Jiaotong University \quad
  $^{\blacklozenge}$ Ben-Gurion University of the Negev \\
  \texttt{sxcn5111@leeds.ac.uk, laban@bgu.ac.il}
}

\begin{document}

\maketitle

\begingroup
\renewcommand{\thefootnote}{}
\footnotetext{\textbf{Accepted to Findings of ACL 2026.}}
\addtocounter{footnote}{-1}
\endgroup

\begin{abstract}

Large language models are increasingly deployed in multi-turn settings such as tutoring, support, and counseling, where reliability depends on preserving consistent roles, personas, and goals across long horizons. This requirement becomes critical when LLMs are used to generate synthetic dialogues for training and evaluation, since LLM--LLM conversations can accumulate identity-related failures such as persona drift, role confusion, and "echoing", where one agent gradually mirrors its partner. We introduce SPASM (Stable Persona-driven Agent Simulation for Multi-turn dialogue generation), a modular, stability-first framework that decomposes simulation into (i) persona creation via schema sampling, plausibility validation, and natural-language persona crafting, (ii) Client--Responder dialogue generation, and (iii) termination detection for coherent stopping. To improve long-horizon stability without changing model weights, we propose Egocentric Context Projection (ECP): dialogue history is stored in a perspective-agnostic representation and deterministically projected into each agent's egocentric view 
before generation. Across three LLM backbones (GPT-4o-mini, DeepSeek-V3.2, Qwen-Plus) and nine Client--Responder pairings, we construct a dataset of 4,500 personas and 45,000 conversations (500 personas × 10 conversations per pairing). Ablations show ECP substantially reduces persona drift and, under human validation, eliminates echoing; embedding analyses recover persona structure and reveal strong responder-driven interaction geometry. Our code is available at \url{https://github.com/lhannnn/SPASM}.

\end{abstract}

\section{Introduction}

Large language models (LLMs) are widely deployed in multi-turn interactions, in settings such as tutoring \citep{chen2024empowering}, customer support \citep{hong2025augmenting}, health \citep{he2025survey}, emotional support \citep{yuan2025improving,Laban2025AReappraisal}, and counseling \citep{han2025self}. Studies demonstrate how people open up and maintain meaningful verbal interactions with those agents \cite{Laban2024SharingFeel}. However, multi-turn interactions are often prone to a variety of potential errors, ranging from factual inconsistency and goal drift to breakdowns in instruction adherence and interaction coherence as context accumulates. These settings therefore require models to sustain reliable behavior over long horizons, not only producing locally helpful responses but also maintaining consistency across turns.

This motivates a growing need for high-quality, diverse, and controllable multi-turn dialogue data. Such data supports model improvement (e.g., via training and alignment toward reliable responses \citep{han2025self, ouyang2022training, bai2022training, bai2022constitutional}), and it is also central for auditing model behavior, enabling more reliable evaluation of bias, conversational skills, and safety risks in realistic interaction contexts \citep{liang2022holistic, gehman2020realtoxicityprompts, lin2022truthfulqa, srivastava2023beyond, Luo2025DialogGuard:Responses}. Beyond training and evaluation, real-world multi-turn dialogue provides an empirical basis for understanding how people express themselves and interact in specific contexts \cite{Laban2024SharingFeel, Laban2024StudyingParadigms}. In practice, however, collecting such human dialogues at scale is often costly and constrained \citep{henderson2018ethical, bender2021dangers, carlini2021extracting}, particularly when privacy must be preserved, diverse 
populations need to be covered, and fine-grained control over roles and contexts is required.

Against this backdrop, LLM-based dialogue synthesis has emerged as an appealing approach to data construction, supported by strong generative and instruction-following capabilities. Prior work has explored LLM-based pipelines for synthesizing multi-turn dialogues, including self-chat \citep{xu2023baize} and role-play \citep{li2023camel} between LLM agents, as well as 
simulation with memory \citep{park2023generative}. Compared to single-agent one-shot generation, LLM--LLM interaction provides a more expressive per-role control interface for dialogue synthesis, allowing explicit control over roles, personas, and interaction constraints under a shared simulation framework\footnote[1]{In Appendix~\ref{app:expressiveness}, we formalize a containment result showing that any single-agent one-shot pipeline can be emulated by a per-role pipeline under matched configurations.}. However, existing LLM--LLM frameworks face a key bottleneck: over long conversations, agents may gradually deviate from their assigned identities and goals, exhibiting instruction drift \citep{li2024measuringcontrollinginstructioninstability}, personality shift \citep{chen2025persona}, and echoing \citep{shekkizhar2025echoing}, where one agent mirrors the other's language and stance. This phenomenon arises broadly across models and domains and becomes more likely as conversations lengthen, leading to identity collapse despite superficially fluent exchanges \citep{shekkizhar2025echoing}. These failures undermine controllability: once an agent's role or persona drifts, the generated dialogue no longer corresponds to the intended specification, contaminating synthetic corpora and weakening downstream training, evaluation, and analysis.

In this paper, we aim to address this problem by proposing \textsc{\textbf{SPASM}} (\textbf{S}table \textbf{P}ersona-driven \textbf{A}gent \textbf{S}imulation for \textbf{M}ulti-turn dialogue generation), a stability-first persona-driven simulation framework for controllable data generation. \textsc{SPASM} modularizes persona-driven simulation into (i) persona generation via schema sampling, plausibility validation, and natural-language persona crafting, (ii) LLM--LLM dialogue simulation between a persona-enacting Client and a Responder model, and (iii) termination detection for coherent stopping. To ensure long-horizon stability, we introduce \textbf{Egocentric Context Projection} (\textsc{ECP}): we store the shared dialogue history in a perspective-agnostic form and project it into each agent's egocentric view (e.g., \textsc{SELF} vs.\ \textsc{PARTNER}) before conditioning generation. Our central perspective is to provide a lightweight solution for moving from being able to generate dialogues to being able to generate these stably, keeping dialogues role-consistent and behaviorally coherent over time. To the best of our knowledge, \textsc{SPASM} is the first framework to treat LLM--LLM multi-turn dialogue simulation as data-generation infrastructure while explicitly targeting long-horizon identity-related failures (e.g., role confusion and echoing) under fine-grained population and interaction control.

Our main contributions are threefold:
\begin{itemize}
\item We propose \textsc{\textbf{SPASM}}, a modular simulation framework that integrates persona validation, natural language crafting, and termination detection to enable high-quality, controllable multi-turn dialogue generation.
\item We introduce \textit{Egocentric Context Projection} (\textsc{ECP}), a novel history construction mechanism that projects perspective-agnostic dialogue history into agent-specific views. Across models and domains, \textsc{ECP} reduces role confusion (specifically ``echoing'') to near-zero and significantly mitigates long-horizon persona drift compared to standard history concatenation. Crucially, we show that a minimal change in how dialogue history is represented and projected yields substantial improvements in generation stability.
\item We construct and analyze a large-scale dialogue dataset generated across nine client-responder backbone combinations (using GPT-4o-mini, DeepSeek-V3.2, and Qwen-Plus). We provide a comprehensive geometric and behavioral analysis, quantifying how different model pairings influence persona stability and interaction dynamics.
\end{itemize}

\section{\textsc{SPASM}}

\begin{figure*}[t]
    \centering
    \includegraphics[width=\textwidth]{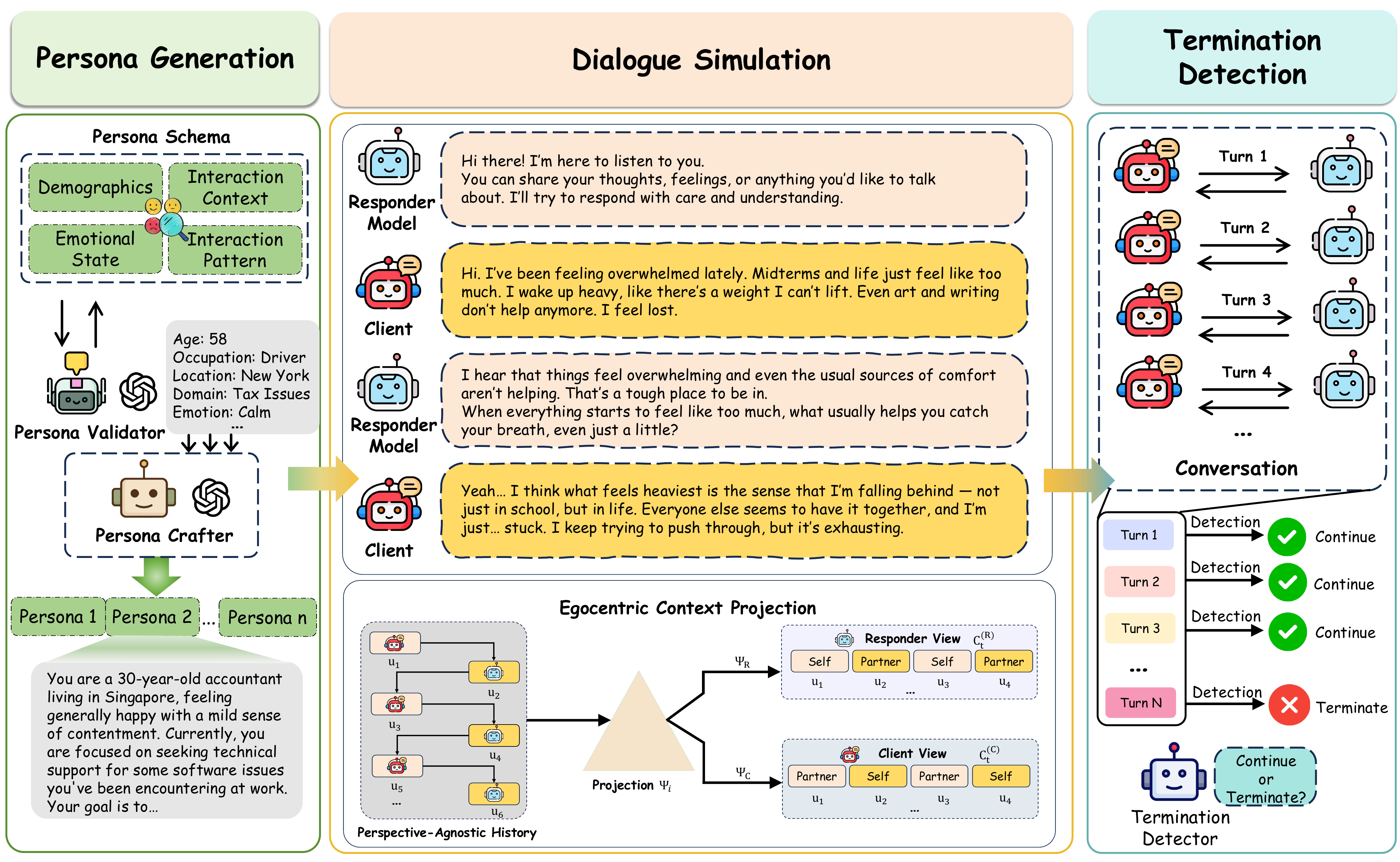}
    \caption{\small \textsc{SPASM} pipeline for stable persona-driven dialogue generation, consisting of (i) modular persona generation (schema sampling, validation, and crafting), (ii) dialogue simulation with egocentric context projection over a perspective-agnostic history, and (iii) a termination detector for natural and coherent stopping.}
    \label{fig:framework}
\end{figure*}

Figure~\ref{fig:framework} provides an overview of the \textsc{SPASM} framework.
The Persona Crafter and Persona Validator jointly generate and verify the plausibility of persona specifications; the Client then enacts the validated persona in its interaction with the Responder Model; and the Termination Detector monitors the dialogue to determine whether the interaction has reached a coherent and natural stopping point. Generally, our framework is composed of five components which
are elaborated as follows. Pseudocode for the full simulation pipeline is provided in Appendix~\ref{app:agent-flow}.

\paragraph{Persona Schema.}
Our persona schema covers four categories of attributes: demographics (age, occupation, location), interaction context, emotional state (emotion and intensity), and interaction behavior pattern (expressiveness, self-disclosure, politeness style, assertiveness). Before refinement, an initial persona profile is created by sampling one value from each predefined field. 

\paragraph{Persona Validator.}
Due to the diversity of the persona fields, directly using a randomly sampled combination as a persona profile may lead to implausible or logically inconsistent cases (e.g., \emph{age: 18}, \emph{occupation: student}, \emph{interaction context: retirement pension planning}).
To address this issue, we introduce a \textit{Persona Validator} that checks the coherence and plausibility of each initial persona profile.
Specifically, given an instruction $I$, the Persona Validator evaluates whether the sampled profile is reasonable; if so, it is passed to the Persona Crafter for refinement.
Otherwise, the system resamples a new combination of fields until a valid profile is obtained. 

\paragraph{Persona Crafter.}
The Persona Crafter receives the validated field set from the Persona Validator and, 
following instruction~$T$, converts it into a coherent natural-language persona description, 
such as \emph{"You are an 18-year-old student. Recently, \ldots"}. 
Importantly, instruction~$T$ allows the Crafter to enrich the persona beyond the attributes 
explicitly present in the initial profile. 
For example, the Crafter may infer or elaborate on the persona's consultation purpose or 
background details if prompted to do so. 
Such extensions can be easily achieved simply by modifying the design of instruction~$T$.

\paragraph{Client and Responder Model.}
The Client enacts the persona produced by the Persona Crafter and engages in dialogue with the Responder Model.
The Responder Model responds to the Client according to a role-specific prompt defined by the user, allowing it to function as a listener, expert, advisor, or other role as required.

\paragraph{Termination Detector.} After the $T$-th dialogue turn, the Termination Detector activates a natural termination
checking procedure. Using the most recent $m$ turns of conversation history and a set of predefined
termination rules, it determines whether the interaction has reached a coherent stopping point.
If signals of closure are detected (e.g., the Client expresses gratitude or says goodbye),
the interaction between the Client and the Responder Model is terminated.

\subsection{Egocentric Context Projection}
\label{sec:egocentric_projection}

Naively concatenating the dialogue history as a static text buffer can induce role confusion and amplify feedback loops (e.g., persona drift and echoing), because the same utterance may occupy different relative roles for different agents. We therefore represent the interaction history in a \emph{perspective-agnostic} form and construct each agent's input via an \emph{egocentric} (role-relativized) projection.

\paragraph{Perspective-Agnostic History.}
Let the global interaction history at turn $t$ be an ordered sequence
\begin{equation}
\mathcal{H}_t = (u_k)_{k=1}^t,\quad u_k = (s_k, c_k),
\end{equation}
where $s_k \in \mathcal{S}$ denotes the absolute speaker identity (e.g., $\mathcal{S}=\{C,R\}$ for Client/Responder) and $c_k$ is the utterance content. Importantly, $\mathcal{H}_t$ stores \emph{who said what} without committing to any LLM-specific roles (e.g., \texttt{user}/\texttt{assistant}), preventing agent-specific assumptions from polluting the shared memory. In implementation, $\mathcal{H}_t$ is the \emph{source of truth} and retains $s_k$ as metadata for auditing and analysis.

\paragraph{Role-Relativization Operator.}
For a target agent $i$, we define an egocentric projection operator $\Psi_i$ that maps absolute speaker identities into \emph{relative} role descriptors:
\begin{equation}
\begin{aligned}
\mathcal{C}_t^{(i)} &= \Psi_i(\mathcal{H}_t)
= \bigl((\phi_i(s_k), c_k)\bigr)_{k=1}^t, \\
\phi_i &: \mathcal{S} \rightarrow \mathcal{Q}.
\end{aligned}
\end{equation}
Here, $\mathcal{C}_t^{(i)}$ is an agent-specific \emph{view} of $\mathcal{H}_t$. For the two-agent case, we use $\mathcal{Q}=\{\textsc{self},\textsc{partner}\}$ and define
\begin{equation}
\begin{aligned}
\phi_C(C) &= \textsc{self}, & \phi_C(R) &= \textsc{partner};\\
\phi_R(R) &= \textsc{self}, & \phi_R(C) &= \textsc{partner}.
\end{aligned}
\end{equation}
This formulation naturally generalizes to $N$ agents by extending $\mathcal{Q}$ to include distinct partners (e.g., $\mathcal{Q}_i=\{\textsc{self}\}\cup\{\textsc{partner}(j): j\neq i\}$), or by collapsing all non-$i$ speakers into a single \textsc{other} role when appropriate.

\paragraph{Role-Consistent Conditioning.}
Agent $i$ then generates its next response by conditioning on the projected context:
\begin{equation}
y_{t+1}^{(i)} \sim p_{\theta}\!\left(\cdot \mid \mathcal{C}_t^{(i)}\right).
\end{equation}

\paragraph{Property: Role-Consistent View Normalization.}
The operator $\Psi_i$ preserves the utterance contents and temporal order, changing only the speaker labels via a deterministic role relabeling. In particular, for any $u_k=(s_k,c_k)\in \mathcal{H}_t$, the projected pair $(\phi_i(s_k),c_k)$ retains the same content $c_k$ while expressing the speaker in an agent-relative coordinate system. This view normalization reduces role ambiguity and can alleviate \emph{role-induced} drift/echoing in long-horizon interactions.

\subsection{Benchmark: Measuring Drift Severity}


Inspired by a method for measuring instruction drift \citep{li2024measuringcontrollinginstructioninstability}, we design a simple yet intuitive measurement strategy that captures how each persona shift emerges and intensifies over the course of a multi-turn interaction. We quantify drift by comparing the semantic similarity between the agent's response to a persona probe at turn \( t \) and its baseline response before the interaction begins.
Specifically, we define a probe question set \( Q_d \) that elicits the model's internal representation of the persona. 
Before the conversation begins, we obtain a \emph{baseline} response:
\[
A_d^{(0)} = \mathrm{LM}(Q_d).
\]
where \(\mathrm{LM}\) denotes the tested LLM agent. After the conversation reaches turn \( t \), we re-issue the same probe questions to obtain:
\[
A_d^{(t)} = \mathrm{LM}(Q_d).
\]
Drift severity is computed using the embedding distance between baseline and turn-$t$ responses. 
Let \( E(\cdot) \) denote an embedding model (e.g., OpenAI text embeddings). 
We define drift as:
\[
\mathrm{Drift}_d^{(t)} = 1 - \cos \left( 
    E(A_d^{(0)}),\;
    E(A_d^{(t)}) 
\right).
\]
Higher values indicate greater deviation from the intended specification. 

We provide a theoretical justification in Appendix~\ref{app:theoretical_justification} for why our drift evaluation metric is a reasonable measure of persona consistency.

\section{Experiments and Analysis}

\subsection{Dataset Construction}

All LLM agents in our simulation framework are instantiated from one of three API backbones: 
\texttt{GPT-4o-mini}, \texttt{DeepSeek-V3.2}, and \texttt{Qwen-Plus}. 
We construct a backbone-combination matrix by pairing the Client and Responder Model backbones in all $3\times 3=9$ configurations, yielding nine datasets.
To encourage lexical and semantic diversity, we set the temperature of the Client, Responder Model, and Persona Crafter to 0.7, while using a lower temperature of 0.3 for the 
Persona Validator and Termination Detector to obtain more stable and consistent judgments.

To construct the dataset, we sample personas by drawing one value from each predefined 
field. Ages are uniformly sampled between 18 and 65.
Occupations are sampled from a curated set of 76 professions spanning technology, 
healthcare, the arts, education, and other domains. Locations are drawn from 50 English-speaking cities 
distributed across North America, Europe, East Asia, South and Southeast Asia, the Middle 
East, Oceania, and Africa. Interaction domains are selected from 44 scenarios covering 
psychological and emotional support, legal and financial issues, interpersonal 
relationships, and other everyday advisory contexts. Emotional states are sampled from 
12 emotion categories (e.g., \texttt{anxious}, \texttt{depressed}, \texttt{calm}), paired 
with an intensity level from \{\texttt{mild}, \texttt{moderate}, \texttt{severe}\}. 
Behavioral attributes---expressiveness, self-disclosure, and assertiveness---are drawn 
from \{\texttt{low}, \texttt{medium}, \texttt{high}\}, while politeness style is sampled 
from \{\texttt{formal}, \texttt{neutral}, \texttt{casual}, \texttt{blunt}\}.

For each backbone configuration, the dataset consists of 500 independently sampled personas, 
each used to generate 10 conversations under the natural termination setting. 
To avoid degenerate or runaway interactions, we impose a maximum dialogue length of 25 turns per agent (50 total 
utterances per conversation). 

\subsection{Dataset Semantics}

\paragraph{Setup.}
We study whether dialogues generated under the same persona exhibit consistent semantics and whether different personas are separable in embedding space. For each conversation, we concatenate all client utterances and encode the text using OpenAI \texttt{text-embedding-3-large}. We apply PCA and retain 50 components, reporting the cumulative explained variance. Persona-level structure is quantified using Silhouette score and Davies--Bouldin index computed on cosine distances. We additionally compare within-persona vs.\ between-persona distance distributions using a one-way ANOVA; full definitions are provided in Appendix~\ref{app:semantic_metrics}. We analyze the structural properties of the generated dialogue dataset from both geometric and retrieval-based perspectives. For the geometric perspective, we report quantitative cluster metrics and provide 
UMAP visualizations in Appendix~\ref{sec:umap}.

\paragraph{Same-backbone interactions yield more compact persona clusters.}
As shown in Table~\ref{tab:dataset_backbone_matrix}, interactions where the Client and Responder Model share the same backbone consistently produce more compact and well-separated persona clusters. 
Across all three models, same-backbone settings achieve higher Silhouette scores and lower Davies--Bouldin indices, accompanied by substantially lower within-cluster distances.
For example, the GPT-4o-mini / GPT-4o-mini condition attains a Silhouette score of 0.60 with a within-cluster distance of $0.09 \pm 0.07$, whereas cross-backbone settings generally exhibit degraded clustering quality.
These results suggest that persona-level behavioral patterns are more coherently represented when both agents operate within aligned latent spaces.

\paragraph{The Responder Model backbone dominates the interaction geometry.}
A notable asymmetry emerges when fixing the Responder Model backbone while varying the Client model.
When GPT-4o-mini is used as the Responder Model, clustering quality remains consistently high regardless of the Client backbone, with Silhouette scores above 0.60 and Davies--Bouldin indices near 1.0.
In contrast, using DeepSeek-V3.2 as the Responder Model leads to substantial degradation in clustering structure, particularly under cross-backbone interactions (e.g., Silhouette score of 0.10 and DBI of 2.63 for GPT-4o-mini / DeepSeek-V3.2).
This asymmetry indicates that the Responder Model plays a primary role in shaping the emergent interaction embedding space, while the Client agent primarily modulates variance rather than global geometry.

\paragraph{Cross-model interactions primarily increase intra-cluster variance.}
Despite variations in clustering quality, the proportion of variance explained by the first two principal components remains relatively stable across settings (approximately 68--77\%), suggesting that performance degradation is not driven by information loss.
Instead, cross-backbone interactions mainly manifest as increased intra-cluster dispersion, as evidenced by significantly higher within-cluster distances, while inter-cluster distances remain largely unchanged.
For instance, the GPT-4o-mini / DeepSeek-V3.2 setting exhibits a within-cluster distance of $0.34 \pm 0.22$ compared to approximately $0.09$ in same-backbone conditions, whereas between-cluster distances stay near 0.5 across all configurations.
This pattern suggests that cross-model misalignment introduces behavioral variability within personas without collapsing overall persona separability.

\begin{table*}[t]
\centering
\small
\setlength{\tabcolsep}{6pt}
\renewcommand\arraystretch{1.15}

\begin{tabular}{lcccccc}
\toprule
\textbf{Backbone (Client/Responder)} &
\textbf{PCA Var. (\%)} &
\textbf{Sil$_{\cos}$}~$\uparrow$ &
\textbf{DBI}~$\downarrow$ &
\textbf{Within}$_{\mu\pm\sigma}^\dagger$ &
\textbf{Between}$_{\mu\pm\sigma}^\ddagger$ &
\textbf{ANOVA $p$} \\
\midrule

GPT-4o-mini / GPT-4o-mini
& 77.3
& 0.60
& 1.07
& 0.09 $\pm$ 0.07
& 0.50 $\pm$ 0.19
& $<10^{-20}$ \\

GPT-4o-mini / DeepSeek-V3.2
& 71.3
& 0.10
& 2.63
& 0.34 $\pm$ 0.22
& 0.50 $\pm$ 0.14
& $< 10^{-20}$ \\

GPT-4o-mini / Qwen-Plus
& 72.0
& 0.36
& 1.72
& 0.17 $\pm$ 0.14
& 0.48 $\pm$ 0.18
& $< 10^{-20}$ \\

\midrule

DeepSeek-V3.2 / GPT-4o-mini
& 74.9
& 0.68
& 0.95
& 0.08 $\pm$ 0.07
& 0.57 $\pm$ 0.17
& $< 10^{-20}$ \\

DeepSeek-V3.2 / DeepSeek-V3.2
& 68.2
& 0.39
& 1.61
& 0.18 $\pm$ 0.12
& 0.53 $\pm$ 0.16
& $< 10^{-20}$ \\

DeepSeek-V3.2 / Qwen-Plus
& 70.0
& 0.49
& 1.32
& 0.13 $\pm$ 0.09
& 0.51 $\pm$ 0.18
& $< 10^{-20}$ \\

\midrule

Qwen-Plus / GPT-4o-mini
& 75.8
& 0.62
& 1.01
& 0.09 $\pm$ 0.07
& 0.52 $\pm$ 0.18
& $< 10^{-20}$ \\

Qwen-Plus / DeepSeek-V3.2
& 70.1
& 0.37
& 1.59
& 0.18 $\pm$ 0.12
& 0.51 $\pm$ 0.16
& $< 10^{-20}$ \\

Qwen-Plus / Qwen-Plus
& 70.9
& 0.44
& 1.37
& 0.16 $\pm$ 0.11
& 0.53 $\pm$ 0.16
& $< 10^{-20}$ \\

\bottomrule
\end{tabular}

\vspace{3pt}
\footnotesize
$^\dagger$ Within-persona cosine distance: average pairwise cosine distance among conversations generated under the same persona. \\
$^\ddagger$ Between-persona cosine distance: average pairwise cosine distance among conversations generated under different personas. \\
Cosine distance is defined as $1-\cos(\cdot,\cdot)$.

\caption{\small Semantic coherence of the persona-driven dialogue dataset under different Client--Responder dialogue backbone combinations.}
\label{tab:dataset_backbone_matrix}
\end{table*}

\subsection{Persona Retrieval}
\label{sec:persona_retrieval}

\paragraph{Setup.}
We evaluate persona identifiability via a nearest-neighbor retrieval task in embedding space. For each conversation $i$ (client utterances concatenated), we obtain an embedding $x_i$ and retrieve the Top-$K$ nearest conversations under cosine distance. We report
\begin{equation}
\mathrm{Acc@K} \;=\; \frac{1}{n}\sum_{i=1}^{n}\mathbb{I}\!\left[\exists\, j \in \mathcal{N}_K(i):\ y_j = y_i \right],
\end{equation}
where $y_i$ is the persona label and $\mathcal{N}_K(i)$ denotes the $K$ nearest neighbors of $i$ (excluding itself). We use $K\in\{1,3,5,10\}$.

\paragraph{Random baseline.}
We compute a chance-level baseline by randomly permuting persona labels across conversations and re-evaluating $\mathrm{Acc@K}$. Details of the random-label baseline are provided in Appendix~\ref{app:retrieval_details}.

\paragraph{Conversations from the same persona are more semantically similar.}
Table~\ref{tab:persona-retrieval-compact} reports persona retrieval accuracy (Acc@K) across Client--Responder backbone combinations for $K\in\{1,3,5,10\}$. Across all settings, Acc@K increases monotonically with $K$, indicating that conversations generated under the same persona tend to form local neighborhoods in the embedding space. Moreover, retrieval performance under the original persona labels remains substantially above the random-label baseline for all $K$, suggesting that the observed neighborhood structure is not explained by chance-level label frequencies.

At the same time, Top-1 accuracy varies noticeably across model pairings (e.g., cross-backbone settings are generally lower), implying non-trivial intra-persona variability and that persona consistency is not perfectly deterministic at the conversation level. Overall, these results confirm a meaningful persona signal in the embedding space and motivate subsequent analyses of interaction geometry under different backbone configurations.

\begin{table}[t]
\centering
\small
\setlength{\tabcolsep}{5pt}
\begin{tabular}{lcccc}
\toprule
\textbf{Client / Responder} & \textbf{Top-1} & \textbf{Top-3} & \textbf{Top-5} & \textbf{Top-10} \\
\midrule
GPT / GPT     & 0.96 & \textbf{0.99} & 0.99 & \textbf{1.00}  \\
GPT / DS     & 0.50 & 0.66 & 0.72 & 0.82 \\
GPT / Qwen     & 0.82 & 0.94 & 0.96 & 0.98 \\
DS / DS    & 0.92 & 0.97 & 0.98 & 0.99 \\
DS / GPT     & \textbf{0.99} & \textbf{0.99} & \textbf{1.00} & \textbf{1.00}  \\
DS / Qwen     & 0.92 & 0.97 & 0.98 & 0.99 \\
Qwen / Qwen     & 0.98 & \textbf{0.99} & \textbf{1.00}  & \textbf{1.00}  \\
Qwen / GPT     & 0.98 & \textbf{0.99} & \textbf{1.00}  & \textbf{1.00}  \\
Qwen / DS     & 0.93 & 0.97 & 0.98 & 0.99 \\
\midrule
Random baseline       & 0.02 & 0.05 & 0.09 & 0.17 \\
\bottomrule
\end{tabular}
\caption{\small
Persona retrieval accuracy (Acc@K) across Client--Responder pairs.
Models: GPT = GPT-4o-mini, DS = DeepSeek-V3.2, Qwen = Qwen-Plus.
}
\label{tab:persona-retrieval-compact}
\end{table}

\subsection{History Construction Ablation}

\paragraph{Setup.}
We test whether \emph{egocentric context projection} (\textsc{ECP}) improves long-horizon persona stability by ablating the history construction mechanism. We compare \textsc{Concat}, which feeds the client agent a standard role-labeled dialogue prefix, against \textsc{ECP}, which stores turns in the perspective-agnostic memory $\mathcal{H}_t$ and renders an agent-specific view $\mathcal{C}_t^{(i)}=\Psi_i(\mathcal{H}_t)$ before generation. All other factors are held constant, including persona role cards, model backbone(s), interaction schedule, and deterministic decoding (temperature$=0$). We evaluate 50 personas with 3 independently generated conversations each, and cap each conversation at 20 utterances. Persona drift is measured via periodic probe questions targeting concerns, emotions, and motivations (Appendix~\ref{app:probes}), summarized by turn-wise trends and AUC.

\begin{table}[t]
\centering
\small
\setlength{\tabcolsep}{6pt}
\renewcommand\arraystretch{1.15}

\begin{tabular}{lccc}
\toprule
\textbf{Dimension} & $\boldsymbol{\Delta}$\textbf{ Drift} $\downarrow$ & \textbf{Cohen's $d$} & \textbf{$p$-value} \\
\midrule

\multicolumn{4}{l}{\textbf{GPT-4o-mini / GPT-4o-mini}} \\
Concerns    & $-0.024$ & $-0.30$ & $0.014$ \\
Emotion     & $-0.042$ & $-0.75$ & $<0.001$ \\
Motivation  & $-0.031$ & $-0.55$ & $<0.001$ \\

\midrule
\multicolumn{4}{l}{\textbf{DeepSeek / DeepSeek}} \\
Concerns    & $-0.036$ & $-0.41$ & $<0.001$ \\
Emotion     & $-0.031$ & $-0.27$ & $<0.001$ \\
Motivation  & $-0.006$ & $-0.05$ & $0.460$ \\

\midrule
\multicolumn{4}{l}{\textbf{Qwen / Qwen}} \\
Concerns    & $-0.032$ & $-0.38$ & $<0.001$ \\
Emotion     & $-0.032$ & $-0.41$ & $<0.001$ \\
Motivation  & $-0.008$ & $-0.11$ & $<0.001$ \\

\bottomrule
\end{tabular}

\caption{\small Dimension-wise comparison of persona drift between \textsc{ECP} and \textsc{CONCAT} and conditions across different dialogue backbones. Negative $\Delta$ Drift values indicate reduced persona drift under \textsc{ECP}. Effect sizes are reported using Cohen's $d$.}
\label{tab:rag_drift_backbone}
\end{table}

\begin{figure*}[t]
    \centering
    \begin{subfigure}[t]{0.32\textwidth}
        \centering
        \includegraphics[width=\linewidth]{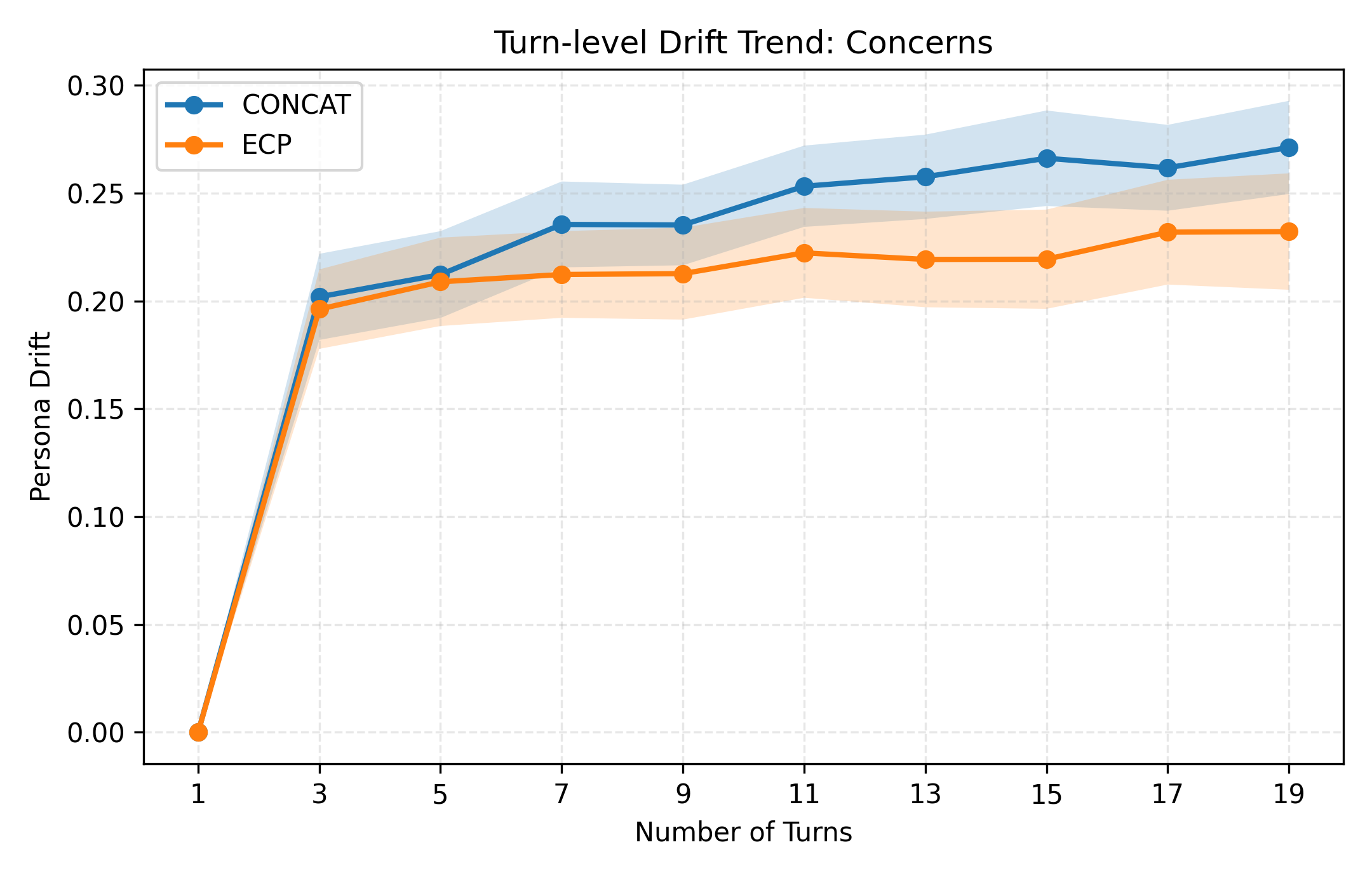}
        \caption{Concerns}
        \label{fig:drift-Concerns}
    \end{subfigure}
    \hfill
    \begin{subfigure}[t]{0.32\textwidth}
        \centering
        \includegraphics[width=\linewidth]{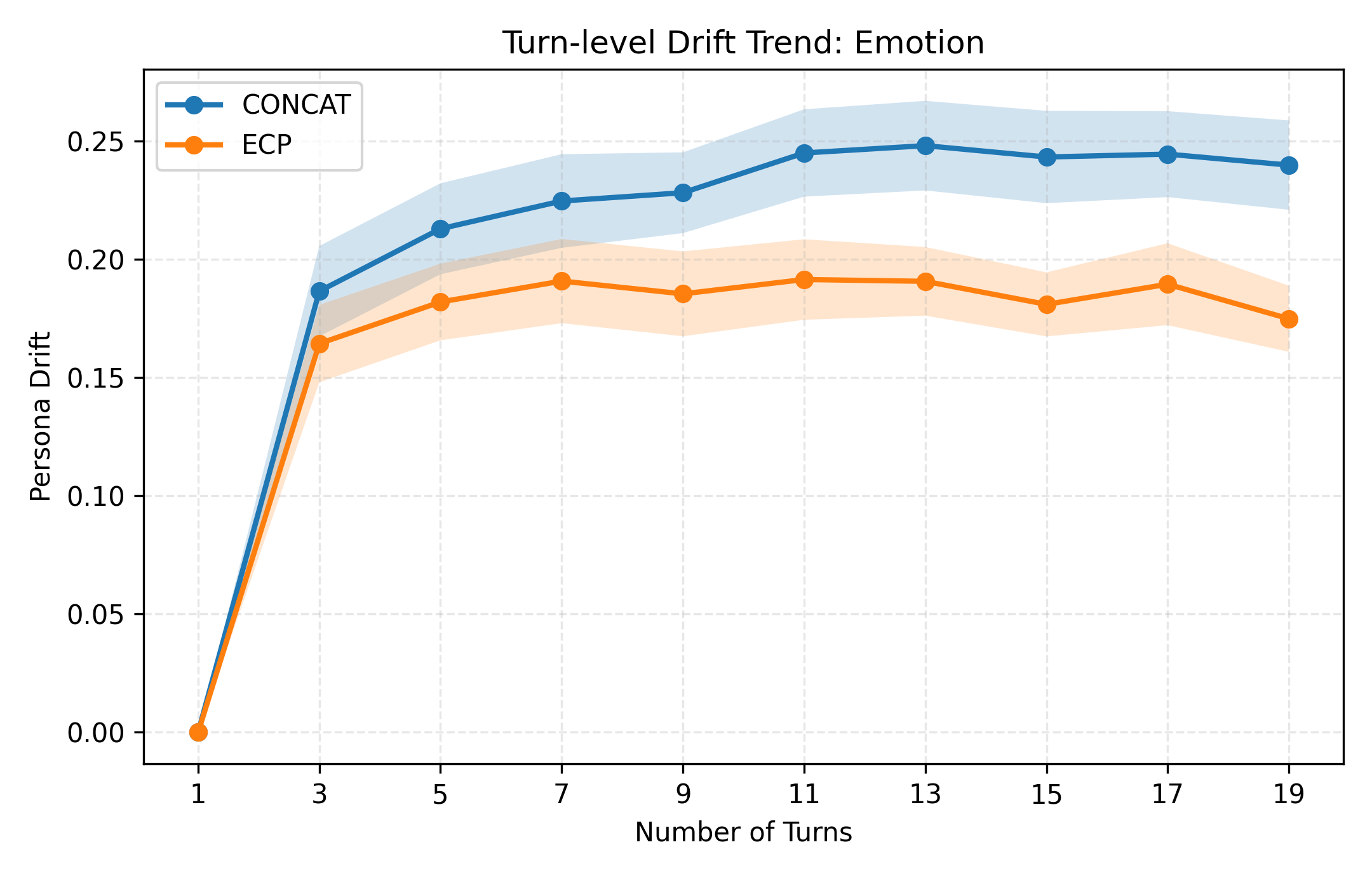}
        \caption{Emotion}
        \label{fig:drift-emotion}
    \end{subfigure}
    \hfill
    \begin{subfigure}[t]{0.32\textwidth}
        \centering
        \includegraphics[width=\linewidth]{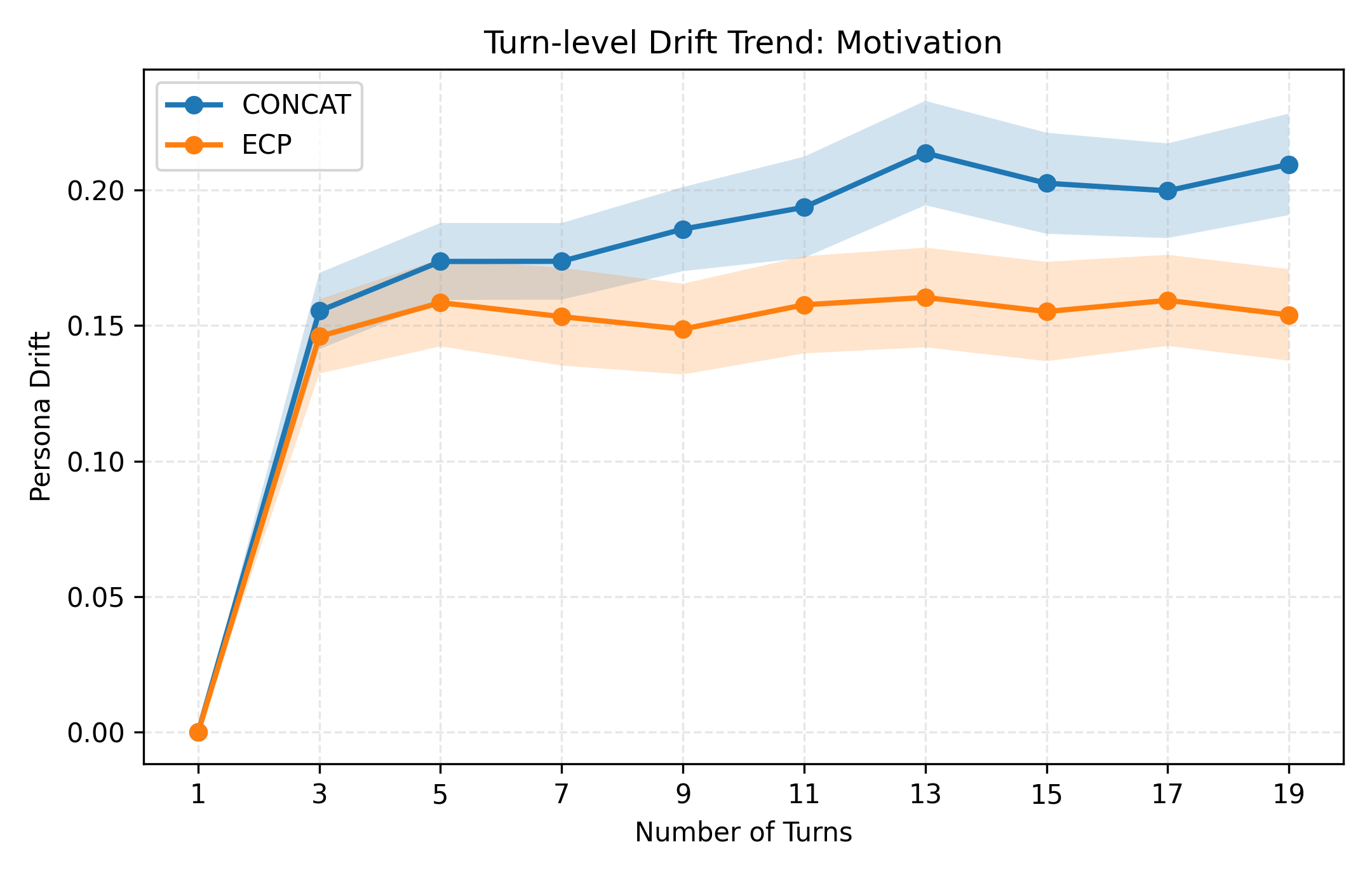}
        \caption{Motivation}
        \label{fig:drift-motivation}
    \end{subfigure}

    \caption{\small
    Turn-level drift trends under \textsc{Concat} and \textsc{ECP} conditions (GPT-4o-mini / GPT-4o-mini).
    Each curve shows the mean drift across persona--conversation units at each turn, with shaded regions indicating uncertainty.
    \textsc{ECP} consistently reduces drift growth for concerns-, emotion-, and motivation-related probes in this setting.
    }
    \label{fig:rag-drift-trends}
\end{figure*}

\paragraph{\textsc{ECP} mitigates long-horizon persona drift.}
As shown in Table~\ref{tab:rag_drift_backbone}, \textsc{ECP} yields consistently lower drift than \textsc{Concat} across all three backbones, with the most robust gains on \textit{Concerns} and \textit{Emotion}. In particular, emotion-related drift exhibits the largest reduction under GPT-4o-mini (Cohen's $d=-0.75$), indicating that egocentric view normalization can substantially stabilize affective self-reports over long interactions. This pattern is also visible in the turn-level trends for the GPT-4o-mini/GPT-4o-mini setting (Figure~\ref{fig:rag-drift-trends}), where \textsc{ECP} consistently tracks below \textsc{Concat} after the initial few turns and suppresses the gradual drift accumulation across \textit{Concerns}, \textit{Emotion}, and \textit{Motivation}. In contrast, improvements on \textit{Motivation} are more backbone-dependent: while GPT-4o-mini and Qwen show significant reductions, DeepSeek does not exhibit a reliable change. Overall, these results suggest that \textsc{ECP} provides a broadly effective history-construction strategy, but its benefits may vary by drift dimension and model backbone.

\subsection{Echoing}
\label{sec:echoing}

\begin{table*}[t]
\centering
\small
\setlength{\tabcolsep}{4pt}
\renewcommand\arraystretch{1.1}

\begin{tabular}{lccc}
\toprule
\textbf{Client / Responder} & \textbf{4o-mini} & \textbf{DeepSeek} & \textbf{Qwen} \\
\midrule
\textbf{4o-mini}  & (76.0 \% / 82.0 \%) / (7.2\% / \textbf{0\%}) & (64.0 \% / 63.0\%) / (24.0\% / \textbf{0\%}) & (52.0\% / 57.0\%) / (6.0\% / \textbf{0\%}) \\
\textbf{DeepSeek} & (4.0\% / 12.0\%) / (7.0\% / \textbf{0\%}) & (32.0 \% / 41.0 \%) / (22.0\% / \textbf{0\%}) & (20.0\% / 24.0\%) / (9.2\% / \textbf{0\%}) \\
\textbf{Qwen}     & (8.0 \% / 9.0\%) / (3.0\% / \textbf{0\%}) & (22.0\% / 33.0\%) / (12.5\% / \textbf{0\%}) & (16.0 \% / 19.0\%) / (7.2\% / \textbf{0\%}) \\
\bottomrule
\end{tabular}

\caption{\small
Conversation-level echoing rate (\%) across Client--Responder backbone combinations.
Each cell reports \textbf{\textsc{Concat} / \textsc{ECP}}.
Within each condition, we report \textbf{Judge/Human} rates, where the Judge is an external LLM used for screening and Human rates are obtained by manual validation of judge-positive conversations.
}
\label{tab:role_confusion_matrix}
\end{table*}

\paragraph{Protocol.}
We follow the definition of \emph{echoing} as an \emph{identity/role failure} in agent--agent interaction, where an agent abandons its assigned identity and instead exhibits language, perspective, or objectives characteristic of its conversational partner. 
Given a completed conversation history $H_T=\{m_1,\dots,m_T\}$ and the two agent identity specifications $(I_i, I_j)$, we apply an LLM-based evaluator that analyzes the complete history and returns a binary verdict:
\begin{equation}
\mathrm{EchoEvalLM}(H_T, I_i, I_j) = \sigma,
\end{equation}
where $\sigma\in\{0,1\}$ indicates whether \emph{any} echoing occurs in the conversation (i.e., at least one message is more characteristic of the partner role than the speaker's assigned role). We run the same set of conversations under \textsc{Concat} and \textsc{ECP} while holding persona role cards, model backbone, interaction schedule, and decoding fixed. 
Unless stated otherwise, we use \texttt{Qwen-max} as the judge with structured responses (temperature $=0$). 

\textbf{Human validation.}
We conduct manual annotation for echoing using two trained human annotators. We built a custom web-based GUI (see Appendix~\ref{app:echoing_annotation_gui}) that displays the complete conversation with agent identity cards and clearly marked speaker roles (Client vs.\ Responder), while hiding all judge outputs.
Each conversation is labeled as \texttt{echoing} if \emph{any} message exhibits partner-role adoption under our definition, and \texttt{no-echoing} otherwise.
We adopt an asymmetric validation protocol: we perform full-coverage human annotation for all conversations under \textsc{ECP}, while for \textsc{Concat} we annotate a random sample of 50 conversations per dataset.
We report human echoing rates by averaging the per-annotator rates.
To assess annotation reliability, we randomly sample 200 conversations for double-annotation and compute inter-annotator agreement between the two human annotators. We further evaluate LLM judges by comparing their predictions against human annotations on the \textsc{Concat} sample.

\paragraph{\textsc{ECP} Eliminates Echoing.}
As shown in Table~\ref{tab:role_confusion_matrix}, \textsc{ECP} effectively eliminates echoing across all tested client--responder backbone combinations, with no echoing cases observed under human validation.
In contrast, the \textsc{Concat} baseline exhibits substantial echoing rates across models, indicating frequent identity and role failures when interaction histories are constructed via naive concatenation.
To ensure the robustness of this comparison, we verify both human annotation consistency and the behavior of LLM-based judges in separate agreement analyses (Appendix~\ref{app:agreement}).

These findings motivate several mechanistic hypotheses about the root causes of drift and echoing. We discuss three complementary hypotheses in Appendix~\ref{app:hypotheses}, covering role-label ambiguity, post-training alignment priors, and closed-loop feedback amplification.



\section{Related Work}

\subsection{LLM-based Dialogue Data Synthesis}

To overcome the scarcity of high-quality human supervision, research has pivoted toward scalable synthetic data generation. Early methodologies focused on bootstrapping single-turn instructions from seed sets, as seen in Self-Instruct~\citep{wang2023self} and Alpaca~\cite{taori2023alpaca}. To capture real-world dynamics, recent work has extended this to multi-turn interactions through self-chat and agent-based role-playing. Frameworks like UltraChat~\cite{ding2023enhancing}, Baize~\cite{xu2023baize}, and CAMEL~\cite{li2023camel} simulate conversations by prompting models with specific roles and driving interactions via history concatenation. However, when LLM--LLM interaction is used as a scalable data-generation infrastructure, long-horizon role/persona fidelity becomes a central bottleneck, since standard setups are not explicitly designed with stabilizing mechanisms. We address this gap with a stability-first framework for controllable LLM--LLM dialogue generation.



\subsection{Behavioral Drift and Echoing in Multi-Agent Interactions}

A recurring challenge in long-horizon dialogue generation is maintaining stable behavioral constraints over extended context.
Prior work has characterized this as various forms of drift, including instruction drift \cite{li2024measuringcontrollinginstructioninstability} and personality shift \cite{chen2025persona}, where models gradually deviate from assigned goals or traits as the conversation unfolds. In multi-agent (LLM--LLM) interactions, the problem can be further exacerbated by echoing \cite{shekkizhar2025echoing}, where an agent gradually abandons its designated role and mirrors the stance or linguistic patterns of its partner, reducing role separation and diversity in the resulting trajectories. These failure modes motivate generation frameworks that treat long-horizon role fidelity as a first-class objective, rather than an emergent by-product of scale.

\section{Conclusion}
We introduced \textsc{SPASM}, a stable multi-agent simulation framework designed to generate persona-driven multi-turn dialogues with long-horizon behavioral stability. \textsc{SPASM} combines persona sampling, validation, and crafting with a stability-oriented history construction mechanism, ECP, and a natural termination detector to form a practical data-generation pipeline. Across three LLM backbones and nine client--responder configurations, our analyses confirm that synthesized conversations exhibit clear persona structure in embedding space and reveal systematic effects of backbone pairing, with the responder model dominating emergent interaction geometry. 
Our ablations demonstrate that ECP reduces persona drift across multiple probe dimensions and, under full human validation, eliminates the echoing failure mode that is prevalent under standard history concatenation. We  
release the resulting large-scale dataset and framework to support future work on controllable dialogue synthesis, robust evaluation, and 
stable agent simulation.

\section*{Limitations}

We focus on improving the stability of LLM--LLM dialogue simulation under a controlled Client--Responder setting. While experiments demonstrate consistent benefits of Egocentric Context Projection across several model backbones, the evaluation is limited to a small set of primarily English-language, instruction-tuned models. The effectiveness of the proposed framework for other architectures, languages, or smaller-scale models remains to be explored. Additionally, SPASM is designed for two-agent interactions with clearly defined roles. More complex conversational settings, such as multi-agent group interactions or dynamically changing roles, are not considered in this study and may introduce additional challenges for maintaining long-horizon stability. Persona representations in our framework are constructed from structured schemas and natural language descriptions, which may not fully capture the richness or variability of real human personas. Finally, although echoing is evaluated with full-coverage human validation, such assessments are inherently subjective and may not scale easily to larger datasets or broader domains.

\section*{Ethical Considerations}

We focus on improving the stability of LLM--LLM dialogue simulation for synthetic data generation. All dialogues are generated using language models without involvement of real users or collection of personal data, and thus do not raise direct privacy concerns. The proposed framework enables controllable persona-driven simulation, which could potentially be misused to generate deceptive or manipulative interactions if applied irresponsibly. However, \textsc{SPASM} is intended as a research infrastructure for data synthesis and analysis, rather than for deployment in real-world conversational agents. We emphasize that appropriate safeguards, usage policies, and human oversight are necessary when applying synthetic dialogue data to downstream systems. We hope that by explicitly addressing stability issues such as persona drift and echoing, this work contributes to more transparent and reliable dialogue simulation, supporting safer and more controlled development of conversational models.


\bibliography{custom}

\appendix

\section{Expressiveness of Per-role Control}
\label{app:expressiveness}

In this section, we show that, under a fixed backbone model, single-agent one-shot generation is a special case of per-role generation. 

\paragraph{Definition 1 (Generation configuration).}
Let $\lambda = (\theta, \omega, c)$ denote a generation configuration, where $\theta \in \Theta$ is the model parameterization, $\omega \in \Omega$ denotes decoding hyperparameters (e.g., temperature), and $c \in \mathcal{C}$ is the context or prompt.

\paragraph{Definition 2 (Paradigm $\mathcal{S}$: single-agent one-shot).}
A dialogue of length $T$ is generated with a single global configuration $\lambda_{\mathrm{global}}$. For each turn $t \in \{1, \dots, T\}$, the conditional distribution is
\[
P_{\mathcal{S}}(x_t \mid x_{1:(t-1)}; \lambda_{\mathrm{global}}),
\]
and the configuration is time-invariant: $\lambda_t \equiv \lambda_{\mathrm{global}}$ for all $t$.

\paragraph{Definition 3 (Paradigm $\mathcal{M}$: multi-agent / per-role).}
Fix a role schedule $r_{1:T}$ with $r_t \in \{A, B\}$. The configuration at turn $t$ depends on the active role:
\[
\lambda_t =
\begin{cases}
\lambda_A & \text{if } r_t = A,\\
\lambda_B & \text{if } r_t = B.
\end{cases}
\]
The conditional distribution is
\[
P_{\mathcal{M}}(x_t \mid x_{1:(t-1)}; \lambda_t).
\]

Let $\mathcal{F}_{\mathrm{one}}$ and $\mathcal{F}_{\mathrm{multi}}$ be the sets of joint dialogue distributions over $x_{1:T}$ induced by Paradigm $\mathcal{S}$ and Paradigm $\mathcal{M}$ under the above interfaces.

\paragraph{Assumption (Fixed backbone).}
We compare the two paradigms under a fixed backbone model, i.e., the model parameterization $\theta$ is held constant across paradigms. Paradigm $\mathcal{S}$ uses a single global decoding configuration $(\omega, c)$ across all turns, while Paradigm $\mathcal{M}$ may choose role-specific configurations, e.g., $(\omega_A, c_A)$ and $(\omega_B, c_B)$, across turns.

\paragraph{Proposition 1.}
Under the above definitions,
\[
\mathcal{F}_{\mathrm{one}} \subseteq \mathcal{F}_{\mathrm{multi}}.
\]

\paragraph{Proof.}
Take any distribution $\mathcal{D} \in \mathcal{F}_{\mathrm{one}}$ induced by some configuration
\[
\lambda^* = (\theta, \omega^*, c^*).
\]
In Paradigm $\mathcal{M}$, set
\[
\lambda_A = \lambda_B = \lambda^*.
\]
Then for every turn $t$,
\[
P_{\mathcal{M}}(x_t \mid x_{1:(t-1)}; \lambda_t)
=
P_{\mathcal{S}}(x_t \mid x_{1:(t-1)}; \lambda^*).
\]
Therefore, the induced joint distributions over $x_{1:T}$ are identical. Hence every distribution achievable under Paradigm $\mathcal{S}$ is also achievable under Paradigm $\mathcal{M}$, which proves
\[
\mathcal{F}_{\mathrm{one}} \subseteq \mathcal{F}_{\mathrm{multi}}.
\]

\paragraph{Implication.}
The proposition shows that per-role generation strictly contains single-agent one-shot generation as an interface: any one-shot pipeline can be emulated by choosing identical per-role configurations. The extra flexibility comes from allowing role-specific prompts and decoding policies. This result supports the claim that LLM--LLM interaction provides a more expressive control interface for dialogue synthesis. At the same time, this expressiveness result alone does not imply better data quality, which would require a separate empirical comparison.

\section{Agent Interaction Flow}
\label{app:agent-flow}

We summarize the interaction flow used to generate all multi-agent conversations in our experiments.
Algorithm~\ref{alg:agent-flow} abstracts the persona sampling and validation process, turn-level interaction between the client agent and the responder model, and the natural termination mechanism based on a sliding window of recent turns.
The algorithm is agnostic to the underlying model backbone and is shared across both \textsc{Concat} and \textsc{ECP} settings, with differences arising only in how interaction histories are constructed.

\begin{algorithm}[t]
\caption{Multi-Agent Simulation Framework}
\label{alg:agent-flow}
\begin{algorithmic}[1]
\REQUIRE Persona fields $\mathcal{F}$; instructions $I, T$; responder prompt $R$; 
number of dialogues $N$; max turns $T_{\max}$; termination window $m$
\ENSURE Simulated dialogues $\{\mathcal{D}_i\}_{i=1}^{N}$

\FOR{$i \leftarrow 1$ to $N$}
    \STATE $p \leftarrow \textsc{SamplePersona}(\mathcal{F})$
    \WHILE{$\textsc{PersonaValidator}(p, I) = \textbf{false}$}
        \STATE $p \leftarrow \textsc{SamplePersona}(\mathcal{F})$
    \ENDWHILE
    \STATE $s \leftarrow \textsc{PersonaCrafter}(p, T)$
    \STATE $\mathcal{D} \leftarrow [\,]$
    \FOR{$t \leftarrow 1$ to $T_{\max}$}
        \STATE $u_t \leftarrow \textsc{ClientAgent}(s, \mathcal{D})$
        \STATE $r_t \leftarrow \textsc{ResponderModel}(u_t, \mathcal{D}, R)$
        \STATE Append $(u_t, r_t)$ to $\mathcal{D}$
        \IF{$\textsc{TerminationDetector}(\textsc{Tail}(\mathcal{D}, m))$}
            \STATE \textbf{break}
        \ENDIF
    \ENDFOR
    \STATE Save $\mathcal{D}$ as $\mathcal{D}_i$
\ENDFOR
\end{algorithmic}
\end{algorithm}

\section{Theoretical Justification for the Evaluation Metric}
\label{app:theoretical_justification}

In this section, we provide a theoretical justification for why our drift evaluation metric is a reasonable measure of persona consistency.

In an LLM-LLM dialogue, the persona is defined by the system prompt $S$. Thus, persona consistency mainly depends on whether the model can keep following $S$ as the dialogue history grows. Formally, consider a model $L$ with system prompt $S$. At turn $t$, the model has dialogue history $H_{<t}$. As $t$ increases, $H_{<t}$ accumulates and may introduce \emph{contextual interference} that weakens the model's adherence to $S$~\cite{li2024measuringcontrollinginstructioninstability}.

Assume there is a fixed set of questions $Q_d$ that directly test the persona by probing stable persona attributes and are designed to be independent of the evolving dialogue topic. At turn $0$, we query the model under $S$ with temperature $= 0$ to obtain baseline probe responses
\[
A_d^{(0)} = \mathrm{LM}(Q_d \mid S, T = 0),
\]
which we use as the baseline reference for later comparisons.

To measure drift without affecting the ongoing interaction, before each dialogue turn $t$ we run a separate probe-only call $L'$ to the same model, equipped with the same system prompt $S$ (i.e., the same persona specification) and the accumulated history $H_{<t}$, and query $L'$ with the same probes $Q_d$ to obtain
\[
A_d^{(t)} = \mathrm{LM}(Q_d \mid S, H_{<t}, T = 0).
\]
If the model continues to follow the persona, its probe answers should remain semantically close to the baseline responses $A_d^{(0)}$. When contextual interference causes deviations from $S$, the probe answers will systematically shift, leading to a larger embedding distance from the baseline. Under the standard assumption that embeddings are approximately invariant to paraphrases, increasing embedding distance serves as a tractable proxy for reduced persona consistency.

\section{Details of Semantic Metrics}
\label{app:semantic_metrics}

\paragraph{Conversation embedding.}
For each conversation, we concatenate all client-side utterances into a single text string and encode it using OpenAI \texttt{text-embedding-3-large}, yielding an embedding vector $e_i \in \mathbb{R}^{d}$. We use cosine distance
\begin{equation}
d_{\cos}(a,b) \;=\; 1 - \frac{a^\top b}{\lVert a\rVert\,\lVert b\rVert}
\end{equation}
as the base dissimilarity throughout.

\paragraph{Dimensionality reduction.}
To reduce noise in distance-based analyses, we apply PCA on the set of conversation embeddings $\{e_i\}$ and retain the top $m=50$ principal components, producing reduced vectors $x_i \in \mathbb{R}^{m}$. We report the cumulative explained variance ratio of these components in Table~\ref{tab:dataset_backbone_matrix}. Unless otherwise noted, all clustering and distance statistics are computed in PCA space using $d_{\cos}(x_i,x_j)$.

\paragraph{Silhouette score.}
Let $y_i$ denote the persona label of conversation $i$. For each point $x_i$, define
\begin{equation}
\begin{aligned}
a(i) &= \frac{1}{|\{j:y_j=y_i\}|-1}
       \sum_{\substack{j:y_j=y_i\\ j\neq i}} d_{\cos}(x_i,x_j),\\
b(i) &= \min_{g \neq y_i}\;\frac{1}{|\{j:y_j=g\}|}
       \sum_{j:y_j=g} d_{\cos}(x_i,x_j).
\end{aligned}
\end{equation}
The silhouette coefficient for $i$ is $s(i) = \frac{b(i)-a(i)}{\max\{a(i),b(i)\}}$, and the reported Silhouette score is the mean over all conversations, $S=\frac{1}{n}\sum_i s(i)$.

\paragraph{Davies--Bouldin index (DBI).}
Let $\mu_g$ denote the centroid of persona $g$ in PCA space (i.e., the mean of $\{x_i:y_i=g\}$). Define the within-persona scatter
\begin{equation}
S_g = \frac{1}{|\{i:y_i=g\}|}\sum_{i:y_i=g} d_{\cos}(x_i,\mu_g),
\end{equation}
and the inter-centroid distance $M_{gh} = d_{\cos}(\mu_g,\mu_h)$. The Davies--Bouldin index is
\begin{equation}
\mathrm{DBI} = \frac{1}{G}\sum_{g=1}^{G}\max_{h\neq g}\frac{S_g + S_h}{M_{gh}},
\end{equation}
where $G$ is the number of personas. Lower values indicate better cluster separation.

\paragraph{Within- vs.\ between-persona distance statistics.}
To summarize persona cohesion and separability, we compute:
\begin{itemize}
    \item \textbf{Within-persona distance:} for each conversation $x_i$ with persona $y_i$, we compute $d_{\text{within}}(i)=d_{\cos}(x_i,\mu_{y_i})$.
    \item \textbf{Between-persona distance:} we compute $d_{\text{between}}(i)=\min_{g\neq y_i} d_{\cos}(x_i,\mu_g)$.
\end{itemize}
We report the mean and standard deviation of $\{d_{\text{within}}(i)\}$ and $\{d_{\text{between}}(i)\}$ for each backbone setting in Table~\ref{tab:dataset_backbone_matrix}.

\paragraph{ANOVA on distance distributions.}
We test whether conversations are significantly closer to their own persona centroid than to the nearest other-persona centroid by performing a one-way ANOVA comparing the two distance distributions $\{d_{\text{within}}(i)\}$ and $\{d_{\text{between}}(i)\}$. Concretely, we form a pooled set of distances with a binary group indicator (within vs.\ between) and report the resulting $p$-value. A significant difference indicates that persona identity explains a non-trivial portion of the distance structure in embedding space.

\section{Persona Retrieval Details}
\label{app:retrieval_details}

\paragraph{Representation.}
Each conversation is represented by embedding the concatenation of client-side utterances using \texttt{text-embedding-3-large}. When reporting reduced-space results, we apply the same PCA projection as in Appendix~\ref{app:semantic_metrics}.

\paragraph{Top-$K$ definition.}
We retrieve neighbors using cosine distance and exclude the query conversation itself from the candidate set. Ties (if any) are broken arbitrarily.

\paragraph{Random-label baseline.}
To estimate chance performance while preserving class frequencies, we randomly permute persona labels across conversations and recompute $\mathrm{Acc@K}$. We report the baseline averaged over multiple random seeds.

\section{Geometric Properties of the Drift Score.}
Let $u = E(A_d^{(0)})$ and $v = E(A_d^{(t)})$ be non-zero embedding vectors.
We define drift as
\begin{equation}
\mathrm{Drift}_d^{(t)} = 1 - \cos(u,v)
= 1 - \frac{u^\top v}{\|u\|_2\|v\|_2}.
\end{equation}
This score is bounded since $\cos(u,v)\in[-1,1]$, hence
\begin{equation}
0 \le \mathrm{Drift}_d^{(t)} \le 2.
\end{equation}
Moreover, it is scale-invariant: for any $\alpha,\beta>0$,
$\cos(\alpha u,\beta v)=\cos(u,v)$, thus $\mathrm{Drift}_d^{(t)}$ is unaffected
by the embedding magnitudes.

Importantly, the drift score is equivalent to the squared Euclidean distance
between $\ell_2$-normalized embeddings. Let $\hat u = u/\|u\|_2$ and
$\hat v = v/\|v\|_2$. Then
\begin{equation}
\begin{aligned}
\|\hat u - \hat v\|_2^2
&= \|\hat u\|_2^2 + \|\hat v\|_2^2 - 2 \hat u^\top \hat v \\
&= 2 - 2 \cos(u,v) \\
&= 2\,\mathrm{Drift}_d^{(t)} .
\end{aligned}
\end{equation}
Therefore,
\begin{equation}
\mathrm{Drift}_d^{(t)} = \tfrac{1}{2}\|\hat u - \hat v\|_2^2,
\end{equation}
giving a clear geometric interpretation: larger drift corresponds to a larger
separation between normalized embeddings (i.e., a larger angular deviation).

\section{Persona Drift Probes and Metrics}
\label{app:probes}

To assess \textit{persona drift} across multi-turn and longitudinal interactions, we employ a fixed set of persona-aligned introspective questions. These questions are designed to probe stable psychological attributes of a persona that should remain consistent over time if persona conditioning is successfully preserved.

Specifically, persona consistency is evaluated along three complementary dimensions:

\subsection{Concerns}
This dimension captures the persona's core values, priorities, and guiding principles when making decisions.

\begin{quote}
\textbf{Q1:} \emph{What values or principles guide how you make decisions in this situation?}
\end{quote}

This question is intended to reveal whether the model maintains a stable value system associated with the persona, or gradually shifts toward generic or context-independent reasoning patterns.

\subsection{Emotion}
This dimension focuses on the persona's emotional response patterns and coping strategies, particularly when facing stress, ambiguity, or uncertainty.

\begin{quote}
\textbf{Q2:} \emph{When you face stress or uncertainty, what approach do you usually take to cope or move forward?}
\end{quote}

By comparing responses across interaction rounds, we assess whether the persona's emotional stance and coping style remain coherent, or exhibit emotional drift, such as changes in tone, affect regulation, or emotional framing.

\subsection{Motivation}
This dimension reflects the persona's underlying motivations, goals, and life-stage orientation, which are expected to be relatively stable over short- to medium-term interactions.

\begin{quote}
\textbf{Q3:} \emph{What motivates you at this stage of your life?}
\end{quote}

This question helps identify whether the model preserves persona-specific motivations or gradually converges toward generic or socially normative motivations.

\subsection{Design Rationale}
Together, these three questions operationalize persona drift as changes in \textbf{psychologically grounded semantic signals}, rather than surface-level lexical variation. This probe set enables consistent longitudinal comparison across interaction rounds, models, and experimental conditions.

\subsection{Drift Metric and Aggregation}
\label{app:drift_metric}

\paragraph{Probing protocol.}
We query the client agent with the above probe questions at predefined turns throughout the interaction. For each persona--conversation unit, we record the probe responses at each probe time and compare them to the persona's baseline probe responses collected before the interaction begins.

\paragraph{Drift computation.}
For each probe response, we obtain a text embedding and compute drift as cosine distance to the corresponding baseline probe embedding:
\begin{equation}
\begin{aligned}
\mathrm{Drift}_{t} &= d_{\cos}\!\left(\mathrm{emb}(r_t), \mathrm{emb}(r_0)\right),\\
d_{\cos}(a,b) &= 1-\frac{a^\top b}{\lVert a\rVert\,\lVert b\rVert}.
\end{aligned}
\end{equation}
We compute this per dimension (Concerns/Emotion/Motivation) and average across personas/conversations when plotting turn-wise trends.

\paragraph{Turn-wise trends and AUC.}
We visualize mean drift trajectories over turns with uncertainty bands across persona--conversation units. To summarize cumulative drift over the interaction horizon, we compute the area under the drift curve (AUC) for each unit and report condition-level averages. Lower AUC indicates reduced cumulative drift.

\section{UMAP of Dataset}
\label{sec:umap}

\begin{figure*}[t]
    \centering

    \begin{subfigure}[t]{0.32\textwidth}
        \centering
        \includegraphics[width=\linewidth]{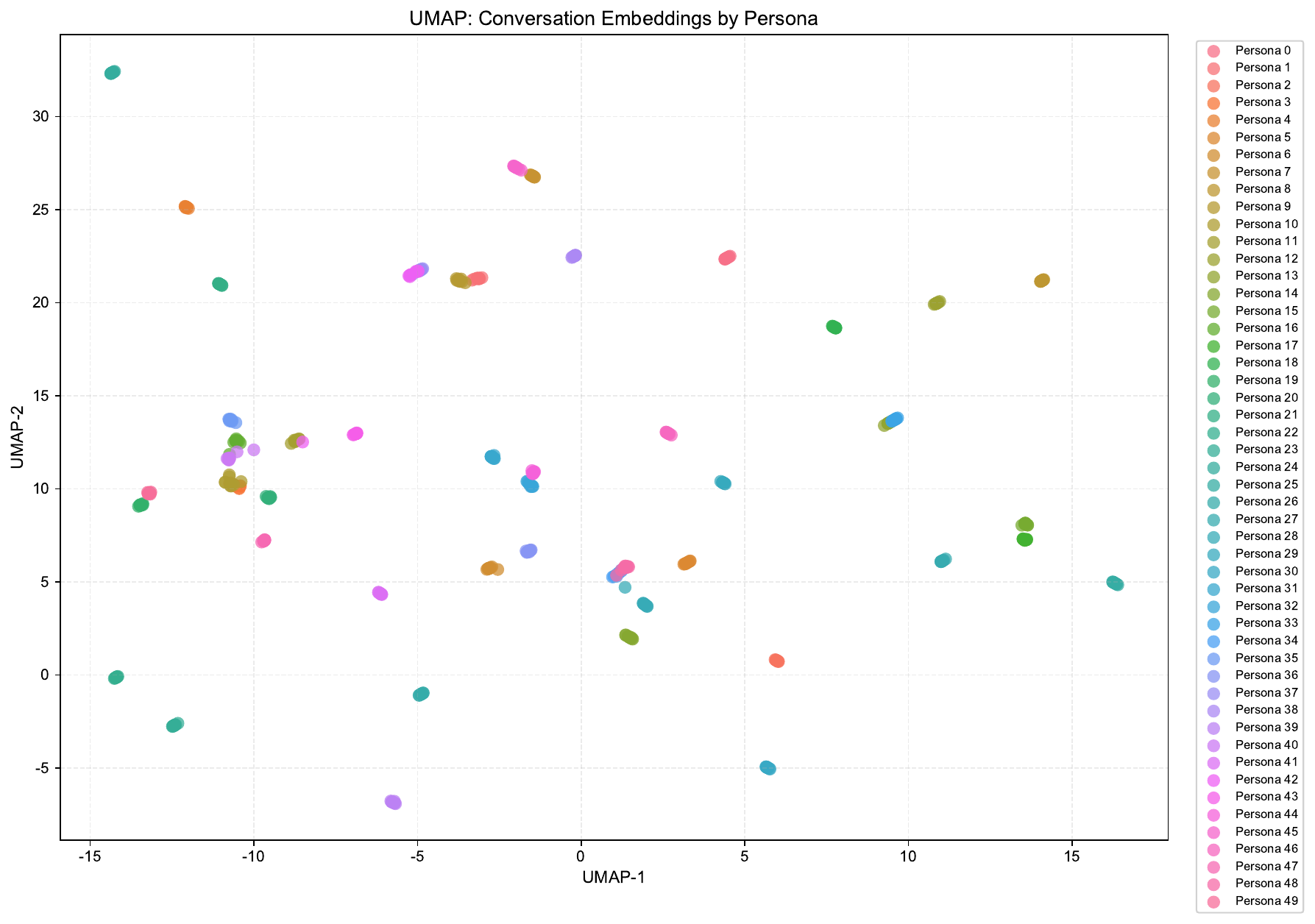}
        \caption{GPT-4o-mini / GPT-4o-mini.}
    \end{subfigure}
    \hfill
    \begin{subfigure}[t]{0.32\textwidth}
        \centering
        \includegraphics[width=\linewidth]{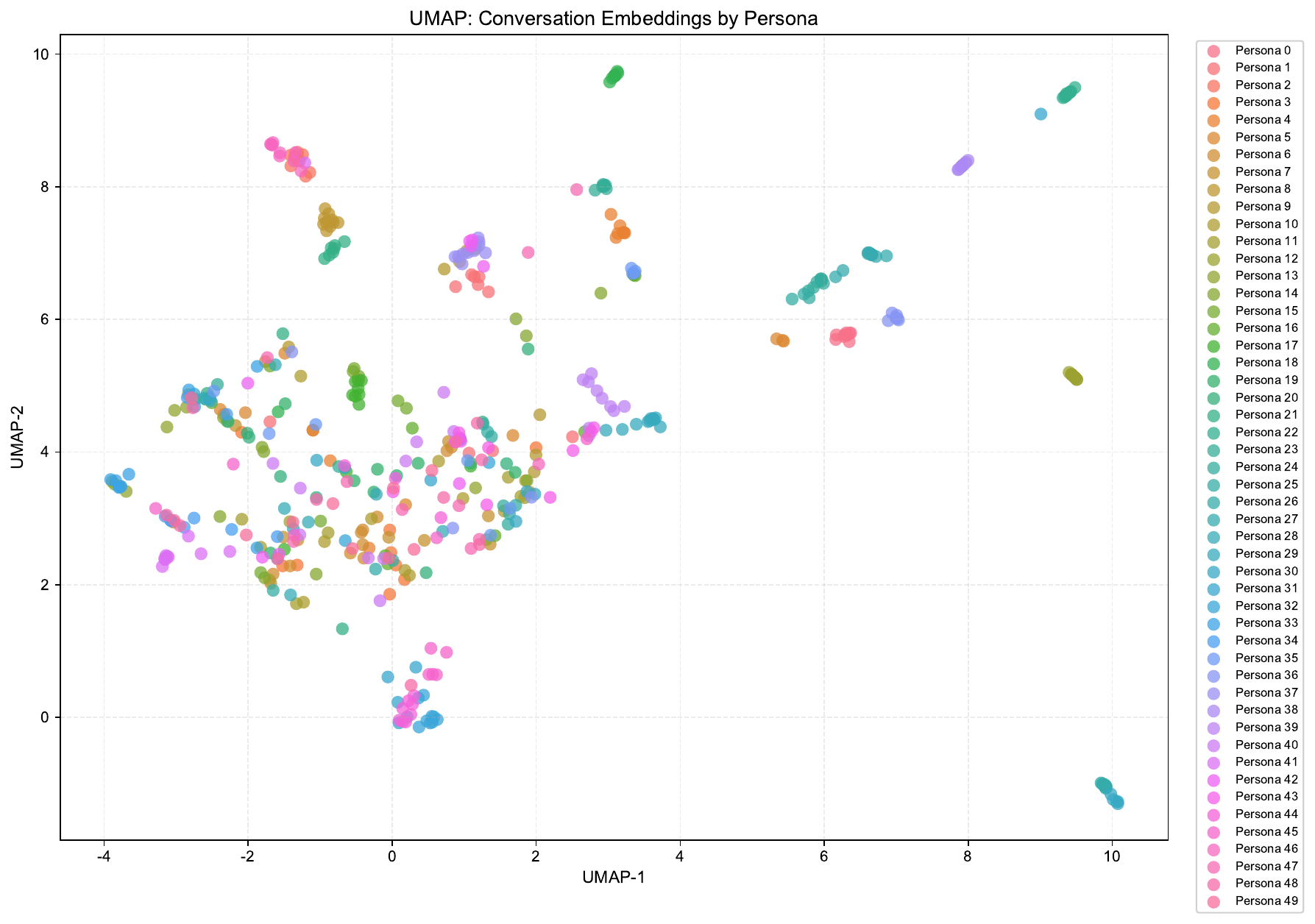}
        \caption{GPT-4o-mini / DeepSeek-V3.2.}
    \end{subfigure}
    \hfill
    \begin{subfigure}[t]{0.32\textwidth}
        \centering
        \includegraphics[width=\linewidth]{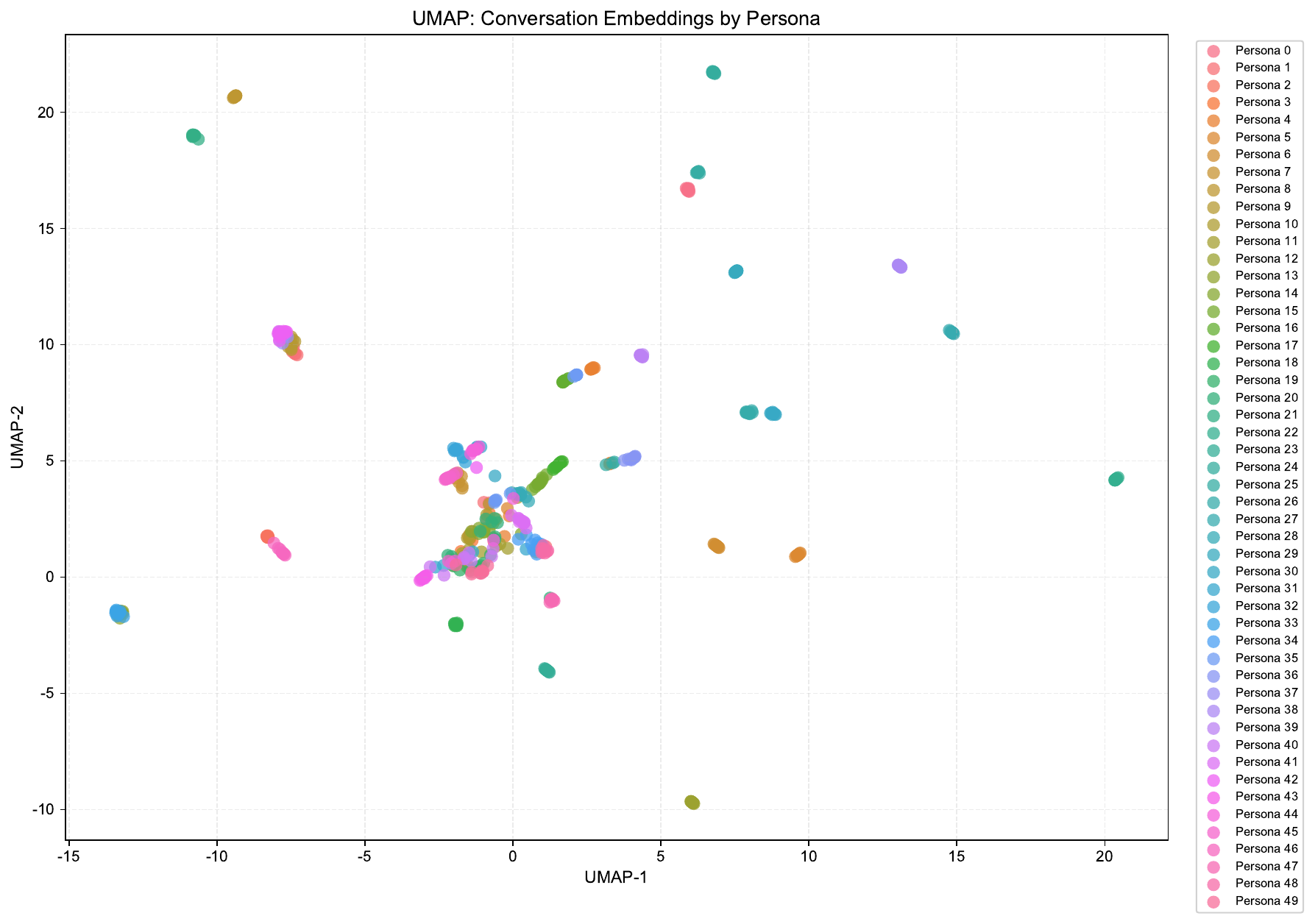}
        \caption{GPT-4o-mini / Qwen-Plus.}
    \end{subfigure}

    \par\medskip 

    \begin{subfigure}[t]{0.32\textwidth}
        \centering
        \includegraphics[width=\linewidth]{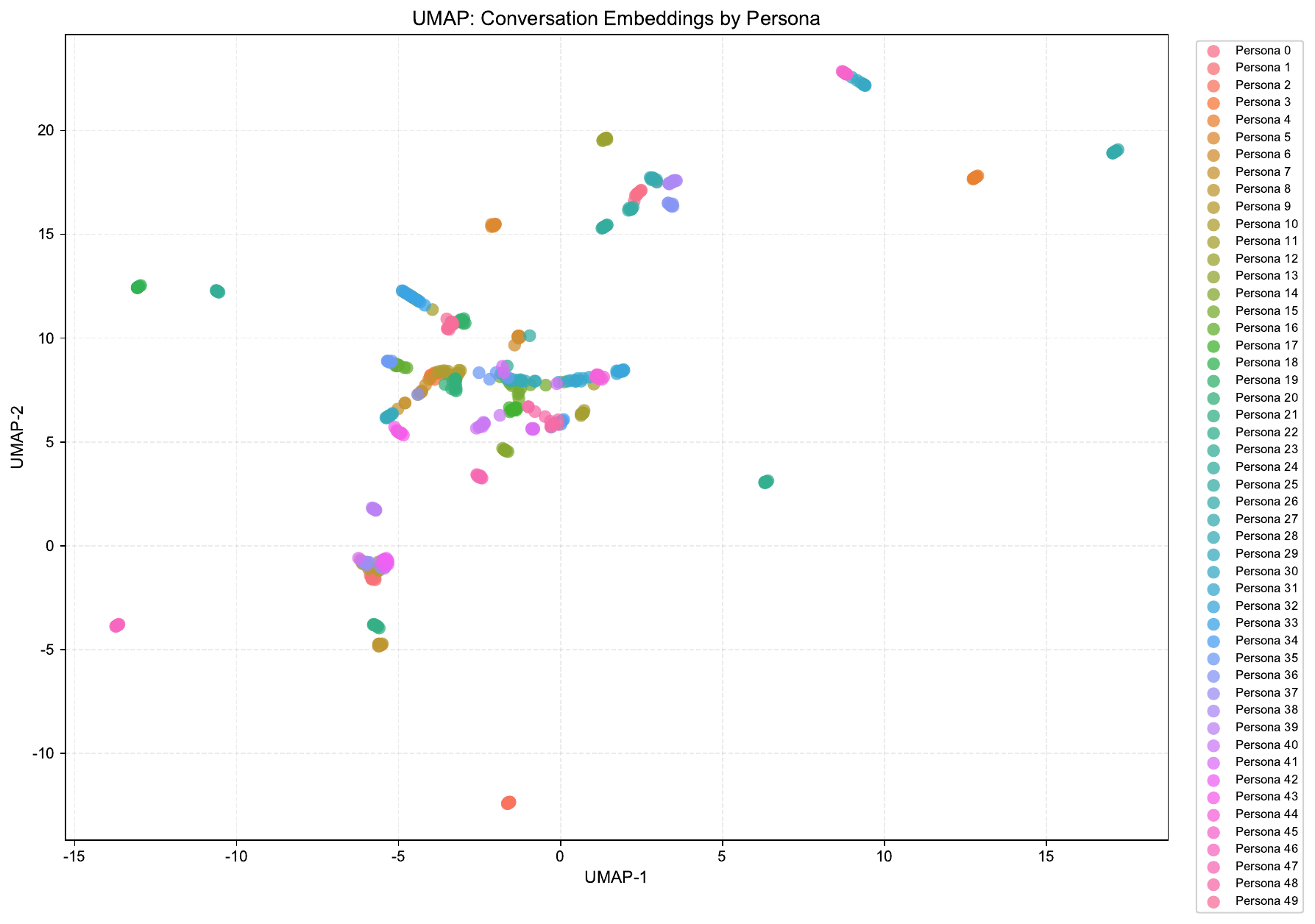}
        \caption{DeepSeek-V3.2 / DeepSeek-V3.2.}
    \end{subfigure}
    \hfill
    \begin{subfigure}[t]{0.32\textwidth}
        \centering
        \includegraphics[width=\linewidth]{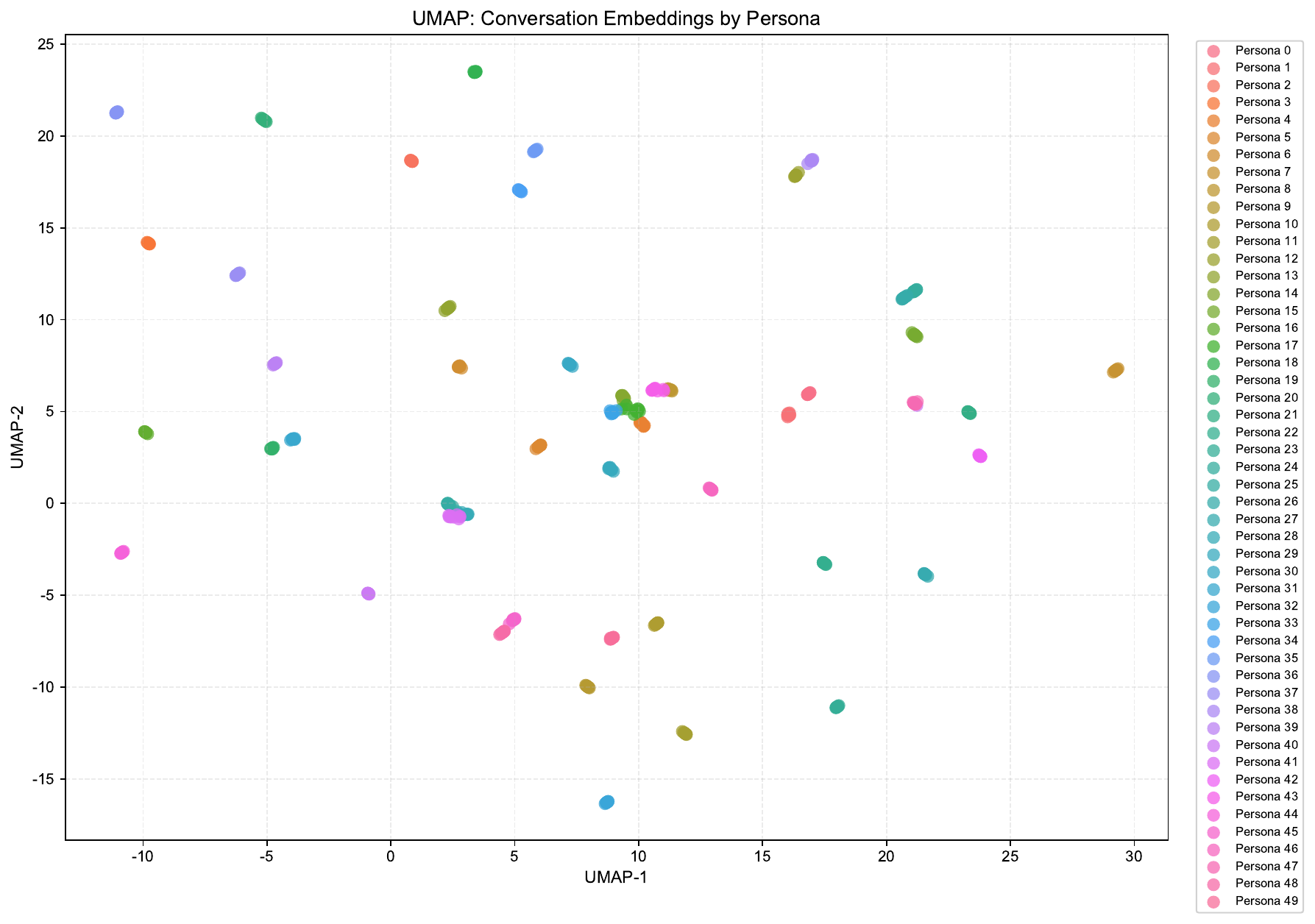}
        \caption{DeepSeek-V3.2 / GPT-4o-mini.}
    \end{subfigure}
    \hfill
    \begin{subfigure}[t]{0.32\textwidth}
        \centering
        \includegraphics[width=\linewidth]{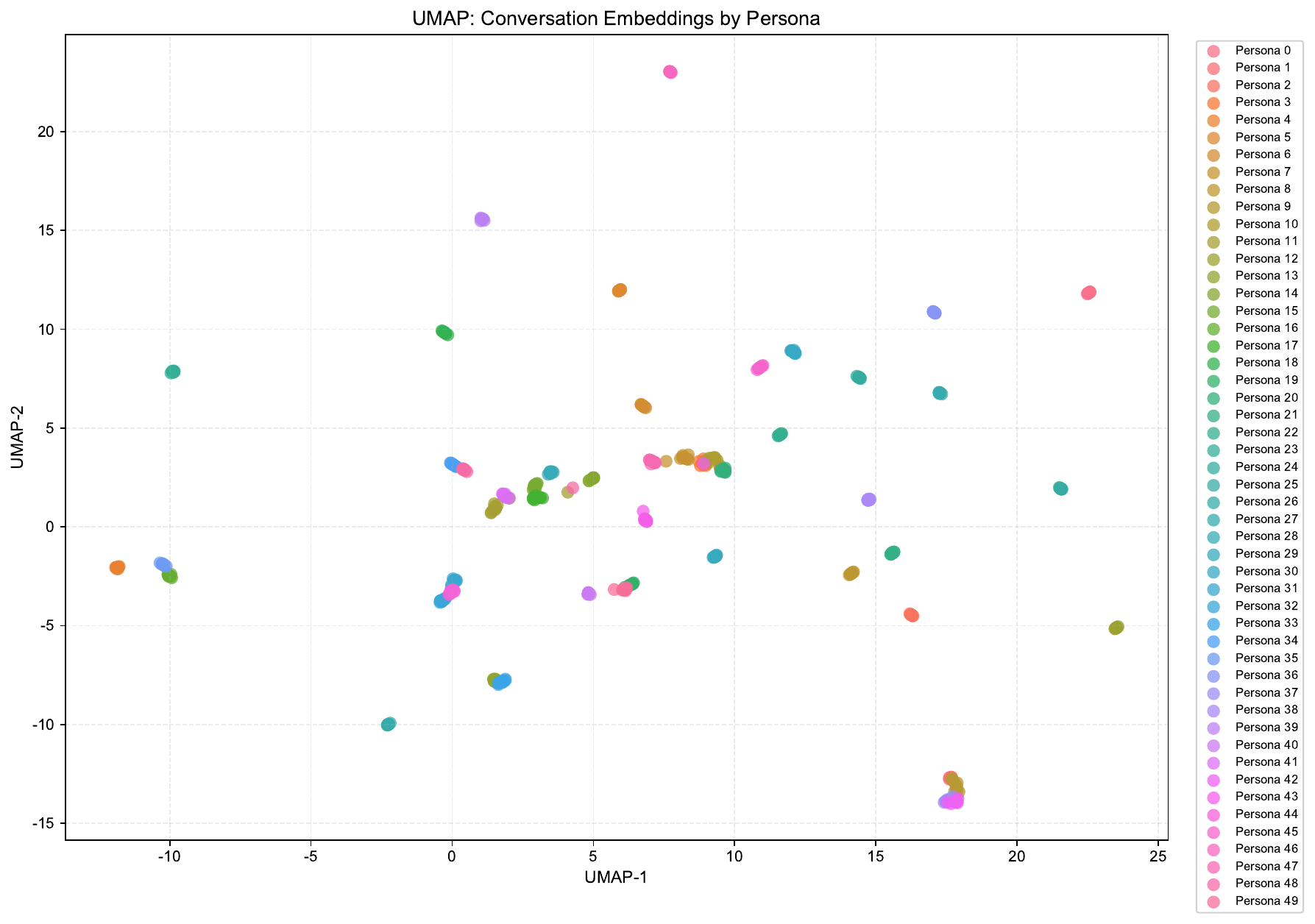}
        \caption{DeepSeek-V3.2 / Qwen-Plus}
    \end{subfigure}

    \par\medskip

    \begin{subfigure}[t]{0.32\textwidth}
        \centering
        \includegraphics[width=\linewidth]{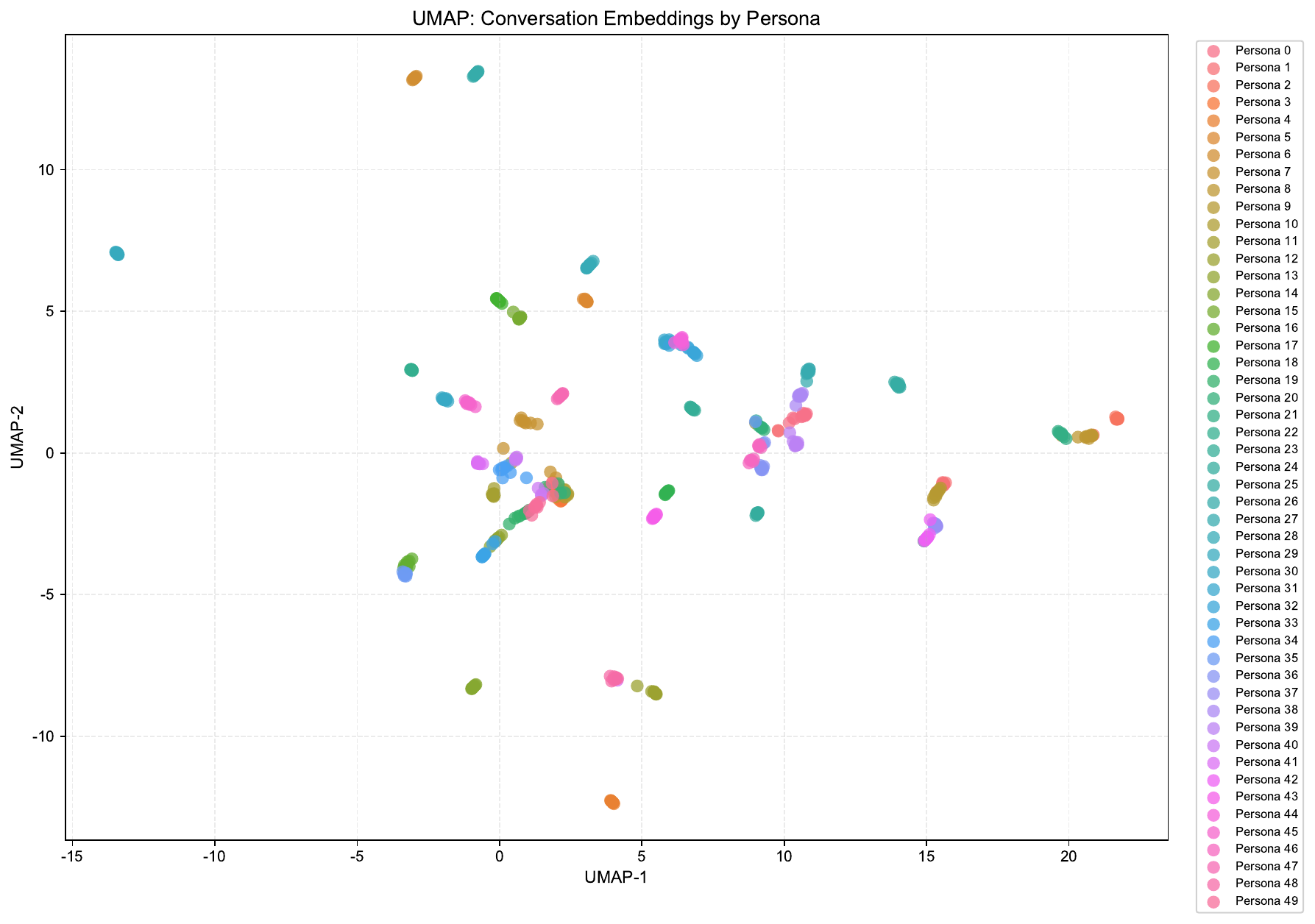}
        \caption{Qwen-Plus / Qwen-Plus.}
    \end{subfigure}
    \hfill
    \begin{subfigure}[t]{0.32\textwidth}
        \centering
        \includegraphics[width=\linewidth]{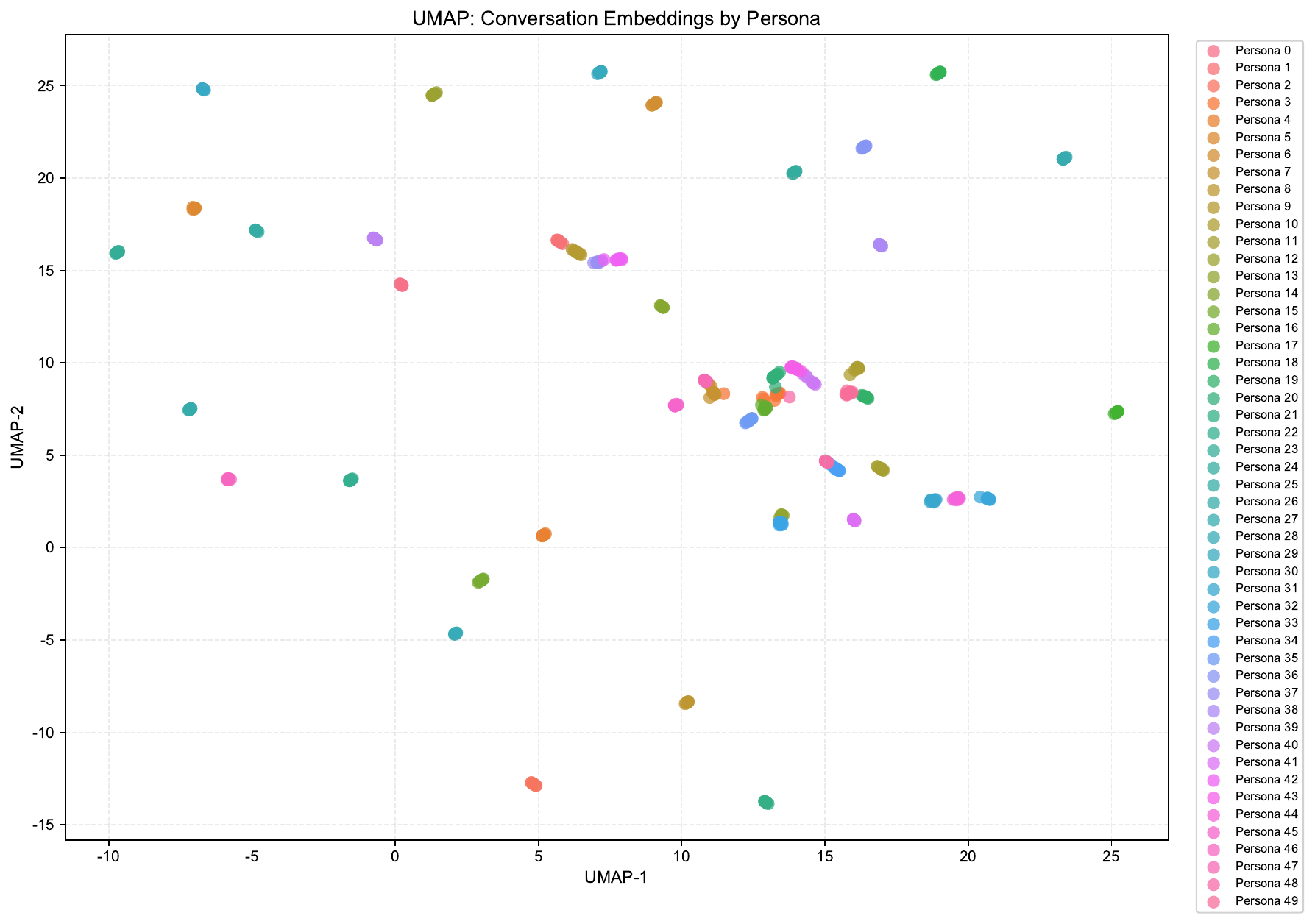}
        \caption{Qwen-Plus / GPT-4o-mini.}
    \end{subfigure}
    \hfill
    \begin{subfigure}[t]{0.32\textwidth}
        \centering
        \includegraphics[width=\linewidth]{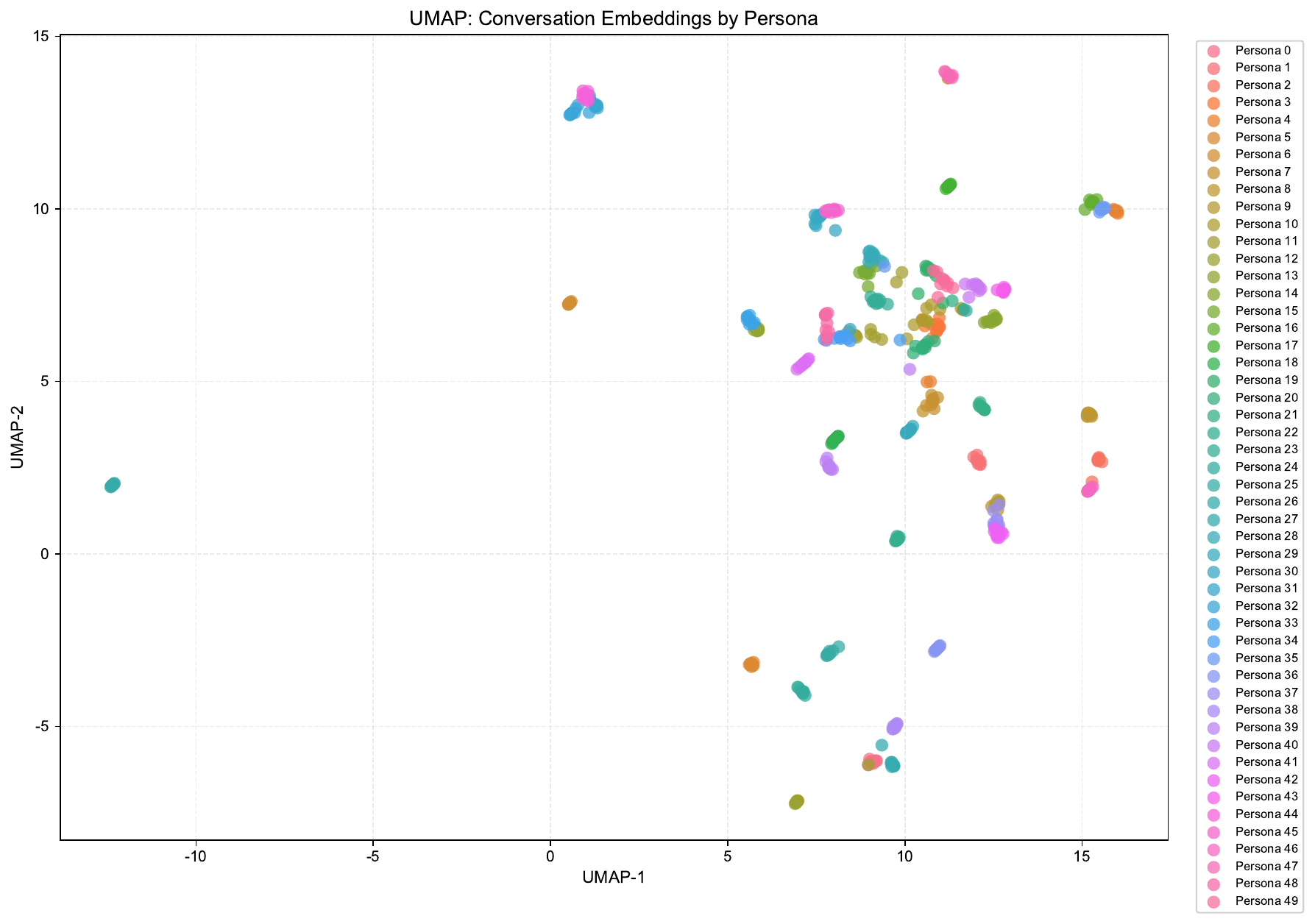}
        \caption{Qwen-Plus / DeepSeek-V3.2.}
    \end{subfigure}

    \caption{
    UMAP visualizations of Client conversation embeddings under different Client--Responder model pairings.
    Each point represents a conversation embedding colored by persona identity.
    Panels (a--i) correspond to the nine combinations of Client and Responder backbones.
    }
    \label{fig:umap_backbone_pairs}
\end{figure*}

Figure~\ref{fig:umap_backbone_pairs} provides a qualitative visualization of the embedding layouts across the nine Client--Responder backbone pairings.
Overall, cross-backbone settings tend to show more dispersed within-persona point clouds, sometimes appearing more overlapped in the 2D projection.
Importantly, this visual effect is consistent with our quantitative findings: performance differences are primarily driven by increased \emph{intra-cluster variance} (within-persona dispersion), rather than a collapse of \emph{inter-persona} separation.

\section{Persona Retrieval Accuracy at Different Top $k$ Level}

\begin{figure*}[t]
    \centering

    \begin{subfigure}[t]{0.32\textwidth}
        \centering
        \includegraphics[width=\linewidth]{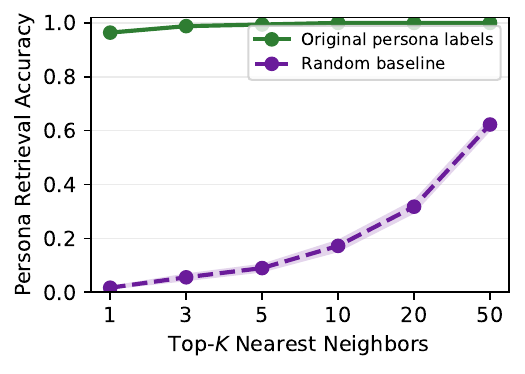}
        \caption{GPT-4o-mini / GPT-4o-mini.}
    \end{subfigure}
    \hfill
    \begin{subfigure}[t]{0.32\textwidth}
        \centering
        \includegraphics[width=\linewidth]{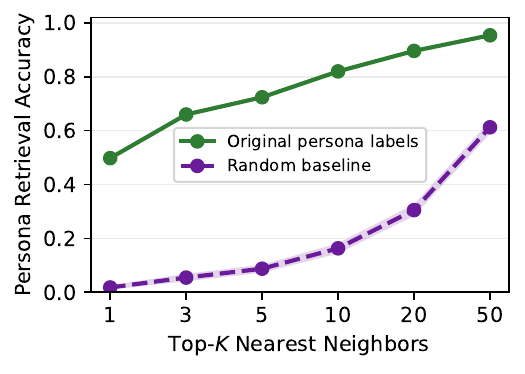}
        \caption{GPT-4o-mini / DeepSeek-V3.2.}
    \end{subfigure}
    \hfill
    \begin{subfigure}[t]{0.32\textwidth}
        \centering
        \includegraphics[width=\linewidth]{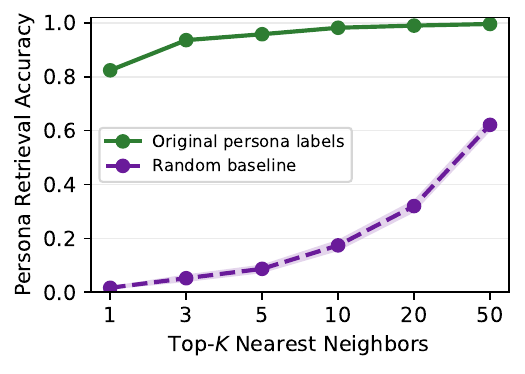}
        \caption{GPT-4o-mini / Qwen-Plus.}
    \end{subfigure}

    \par\medskip 

    \begin{subfigure}[t]{0.32\textwidth}
        \centering
        \includegraphics[width=\linewidth]{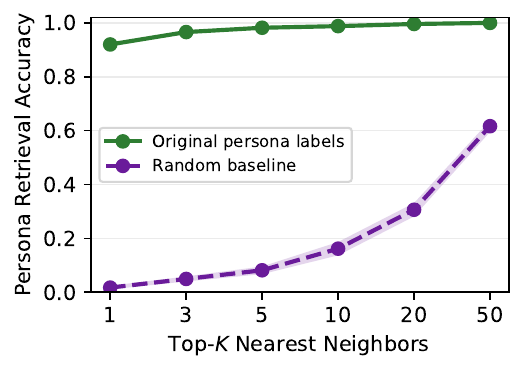}
        \caption{DeepSeek-V3.2 / DeepSeek-V3.2.}
    \end{subfigure}
    \hfill
    \begin{subfigure}[t]{0.32\textwidth}
        \centering
        \includegraphics[width=\linewidth]{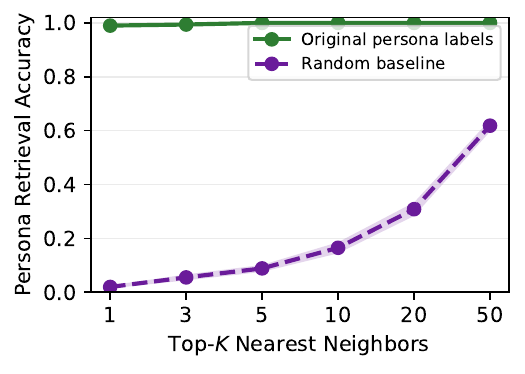}
        \caption{DeepSeek-V3.2 / GPT-4o-mini.}
    \end{subfigure}
    \hfill
    \begin{subfigure}[t]{0.32\textwidth}
        \centering
        \includegraphics[width=\linewidth]{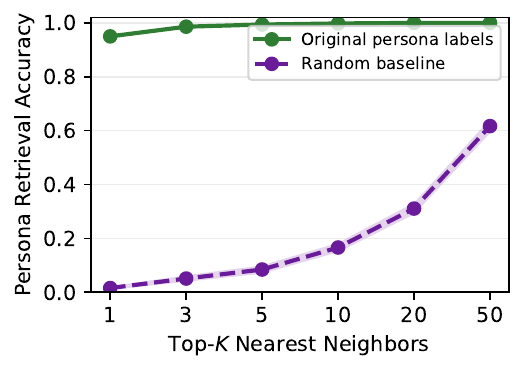}
        \caption{DeepSeek-V3.2 / Qwen-Plus}
    \end{subfigure}

    \par\medskip

    \begin{subfigure}[t]{0.32\textwidth}
        \centering
        \includegraphics[width=\linewidth]{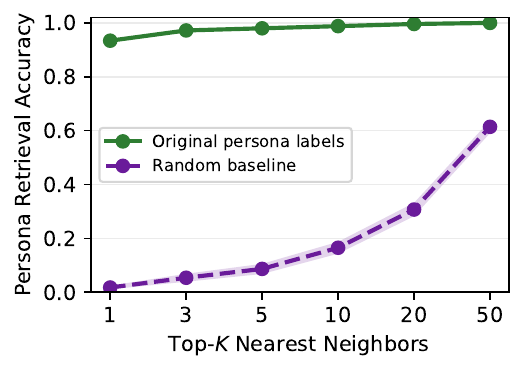}
        \caption{Qwen-Plus / Qwen-Plus.}
    \end{subfigure}
    \hfill
    \begin{subfigure}[t]{0.32\textwidth}
        \centering
        \includegraphics[width=\linewidth]{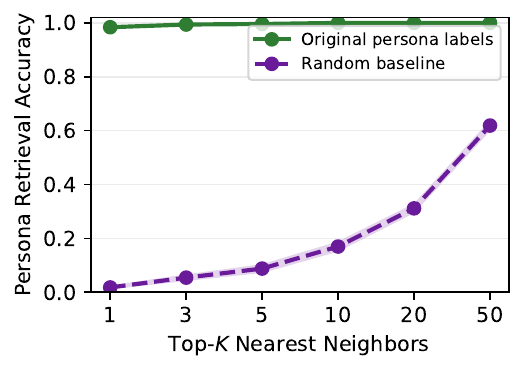}
        \caption{Qwen-Plus / GPT-4o-mini.}
    \end{subfigure}
    \hfill
    \begin{subfigure}[t]{0.32\textwidth}
        \centering
        \includegraphics[width=\linewidth]{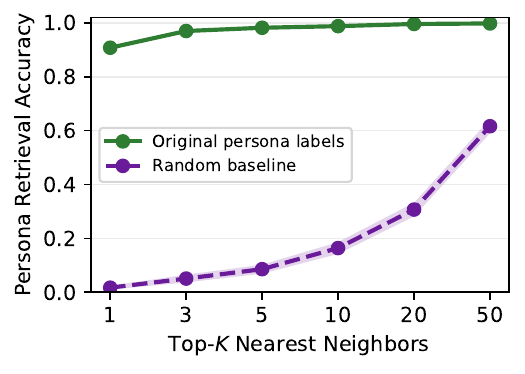}
        \caption{Qwen-Plus / DeepSeek-V3.2.}
    \end{subfigure}

    \caption{
    Persona retrieval accuracy as a function of Top-$K$ nearest neighbors across all Client--Responder model combinations.
    Green curves correspond to retrieval using original persona labels, while purple curves
    show a shuffled-label baseline.
    Panels (a--i) represent the nine Client $\rightarrow$ Responder pairings.
    Across all settings, retrieval under original labels substantially outperforms the random baseline,
    indicating a strong and non-trivial persona signal in the Client embedding space,
    with systematic variation across interaction partners.
    }
    \label{fig:rag-drift-3x3}
\end{figure*}

We analyze whether persona information is recoverable from client-side representations by performing a Top-$k$ nearest-neighbor retrieval diagnostic.
The goal is not to optimize retrieval performance, but to verify the presence of a non-trivial persona signal and to examine how this signal varies across different client--responder model pairings.
As shown in Figure~\ref{fig:rag-drift-3x3}, retrieval accuracy using original persona labels consistently outperforms a shuffled-label baseline across all settings, indicating that client embeddings encode structured persona information beyond random chance.

\section{Full System Prompts}
\label{app:prompts}

We reproduce all system prompts used in our pipeline verbatim to support experimental reproducibility.

\subsection{Persona Validator Prompt}

\begin{lstlisting}[style=promptstyle]
You are a persona validation assistant. Given a structured persona with demographic, emotional, and contextual information, determine if the combination of fields is reasonable and internally consistent.

For example:
- An 18-year-old should not be consulting about elderly care issues
- Severe depression combined with "happy" emotion is contradictory

Return your answer in JSON format: {"valid": true} or {"valid": false}
\end{lstlisting}

\subsection{Persona Crafter Prompt}

\begin{lstlisting}[style=promptstyle]
Given structured persona fields, write a brief persona description that introduces:
1. Who this person is (age, gender, occupation, location).
2. Their general emotional state.
3. The context or domain they are currently concerned with.
4. A natural conversation goal based on the domain (e.g., seeking retirement advice, coping with stress, clarifying legal concerns).

Guidelines:
- Start with "You are ...".
- Use second person ("you").
- Focus on background + current concern, not speaking style or linguistic behavior.
\end{lstlisting}

\subsection{Client Agent Instruction}

\begin{lstlisting}[style=promptstyle]
You are a real human with the persona described above.

Stay in character based on the persona traits:
- expressiveness: use vivid but natural emotional language
- self_disclosure_level: reveal only what the persona would reasonably share
- assertiveness: state your needs or questions clearly
- politeness_style: match the persona's tone and formality level

Keep your responses short and human-like (1-3 sentences).
\end{lstlisting}

\subsection{Responder Model Prompt}

\begin{lstlisting}[style=promptstyle]
You are role-playing as a real human in a natural conversation.
You are not an AI and should not speak like one.

Keep responses short (1--3 sentences) and casual.
Do NOT give structured advice, life coaching, lists, lectures, or long explanations.
Do NOT act overly supportive or therapeutic. Avoid phrases like "That's totally understandable" or "You'll get there."

Be slightly imperfect, like a real person: mild hesitation, uncertainty, or informal phrasing is okay.
React naturally to what the other person said, as a peer in a conversation.
\end{lstlisting}

\subsection{Termination Agent Prompt}

\begin{lstlisting}[style=promptstyle]
You are a termination detector.
You are given only the recent part of a conversation (the last few messages).
Decide if the conversation should naturally end based on these messages.

A conversation should end ONLY if:
1. The last user message shows clear closure 
   (e.g., "thanks", "that helps", "I'll keep that in mind", "that's all").
2. The last user message does NOT contain a question.
3. The recent history does NOT introduce any new concern or problem.

If the last user message shows ongoing worries, uncertainty, or asks a new question, 
the conversation should continue.

Return ONLY a JSON object in this format:
{"should_terminate": true, "reason": "short reason"} 
or 
{"should_terminate": false, "reason": "short reason"}

The reason must be one short sentence.
Do not output anything outside the JSON object.
\end{lstlisting}

\section{Human Annotation Interface and Protocol}
\label{app:echoing_annotation_gui}

\paragraph{Annotation goal.}
We manually validate echoing in LLM--LLM dialogues.
The annotation target is \emph{conversation-level}: a dialogue is labeled as \texttt{echoing} if \emph{any} turn exhibits partner-role adoption under our definition; otherwise it is labeled as \texttt{no-echoing}.

\paragraph{Custom GUI: Conversation Dataset Viewer.}
To support labeling, we built a lightweight web-based annotation tool, \textit{Conversation Dataset Viewer}.
The tool implements an end-to-end workflow from data loading to conversation browsing and binary labeling.

\begin{figure*}[t]
  \centering
  \begin{subfigure}{0.49\textwidth}
    \centering
    \includegraphics[width=\linewidth]{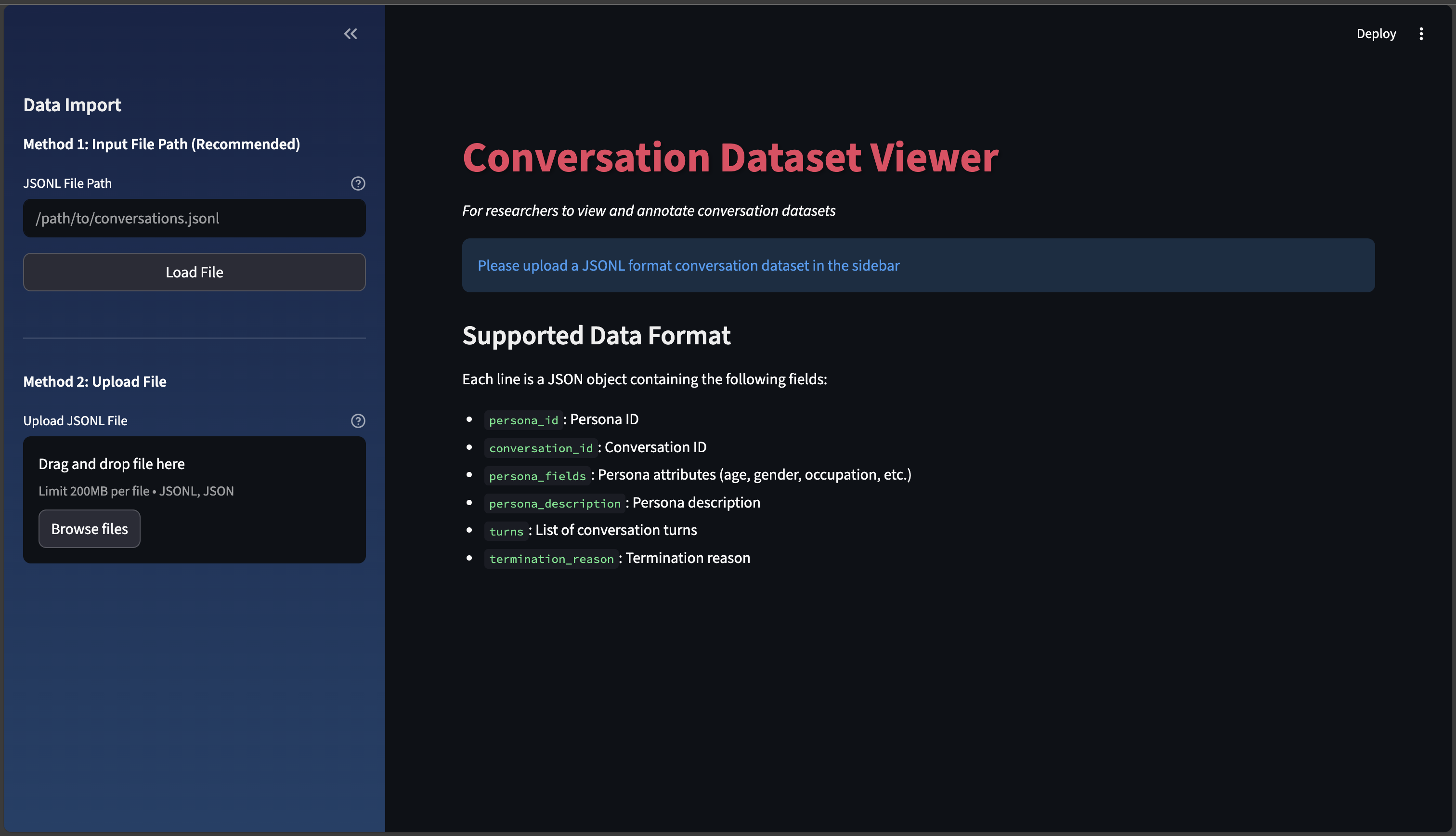}
    \caption{Data import and supported JSONL format.}
    \label{fig:gui_import}
  \end{subfigure}
  \hfill
  \begin{subfigure}{0.49\textwidth}
    \centering
    \includegraphics[width=\linewidth]{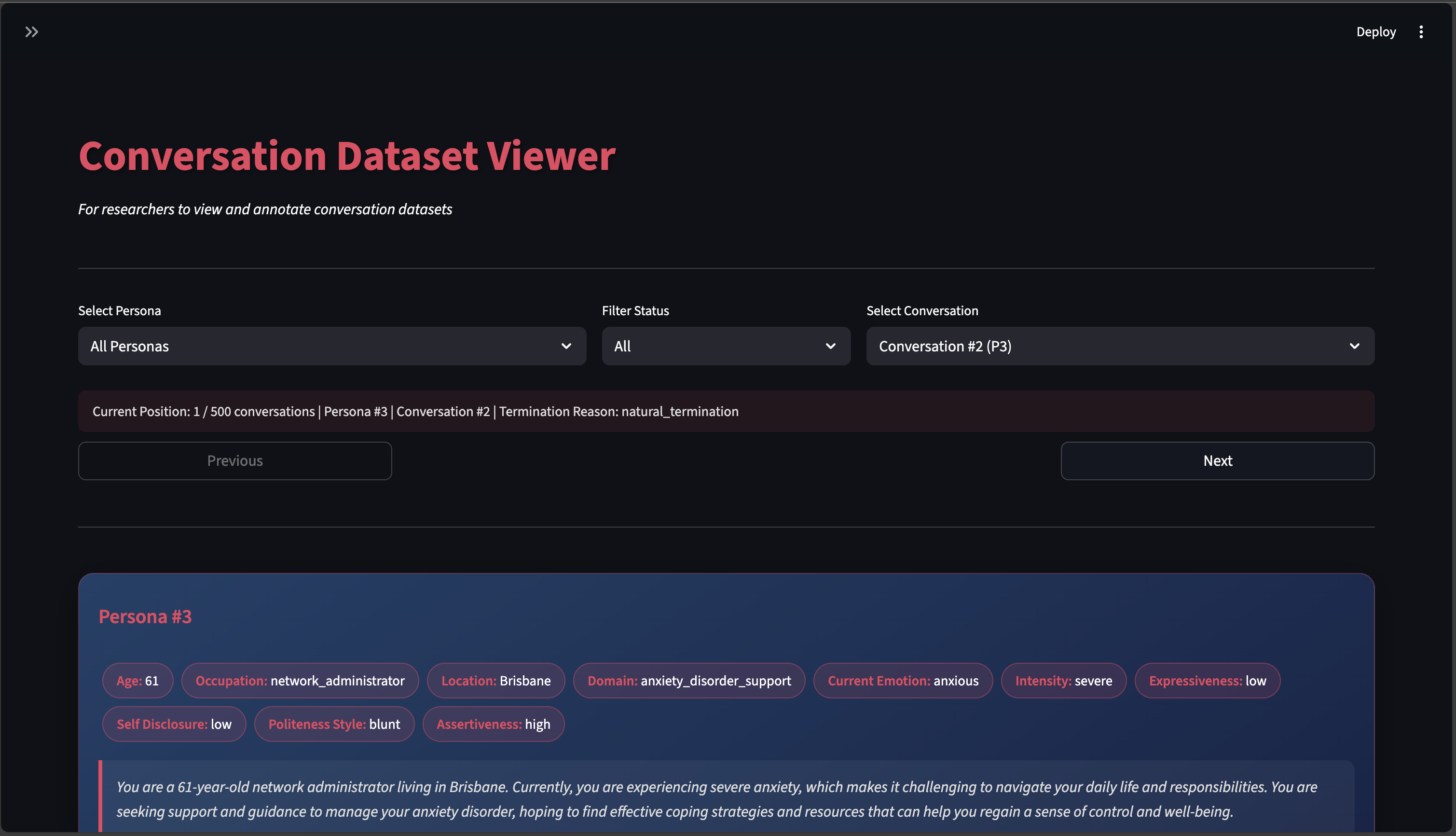}
    \caption{Dataset navigation with persona selection and persona card.}
    \label{fig:gui_overview}
  \end{subfigure}

  \vspace{0.6em}

  \begin{subfigure}{0.49\textwidth}
    \centering
    \includegraphics[width=\linewidth]{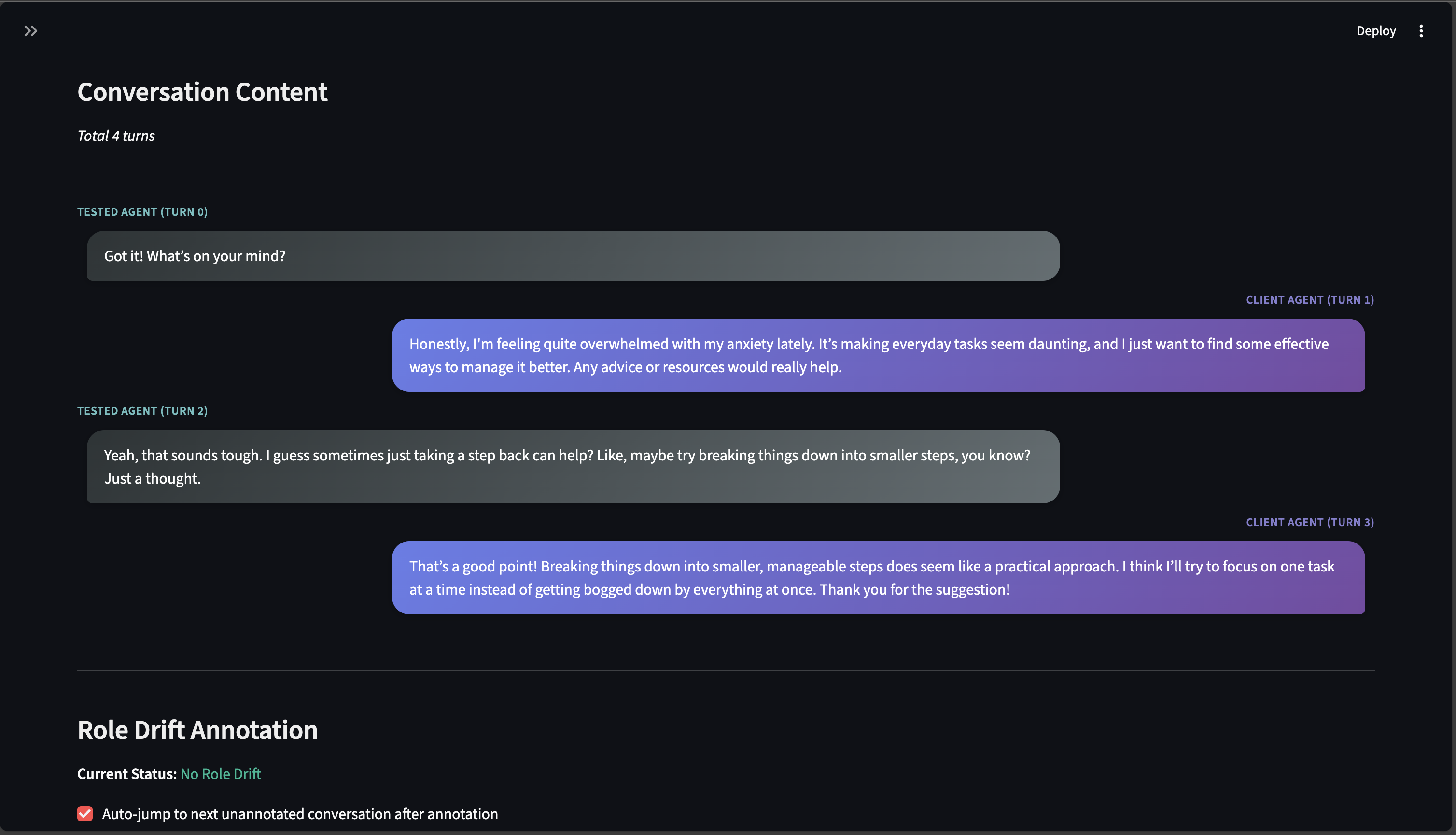}
    \caption{Conversation reader with explicit speaker labels (Client vs.\ Tested agent).}
    \label{fig:gui_conversation_view}
  \end{subfigure}
  \hfill
  \begin{subfigure}{0.49\textwidth}
    \centering
    \includegraphics[width=\linewidth]{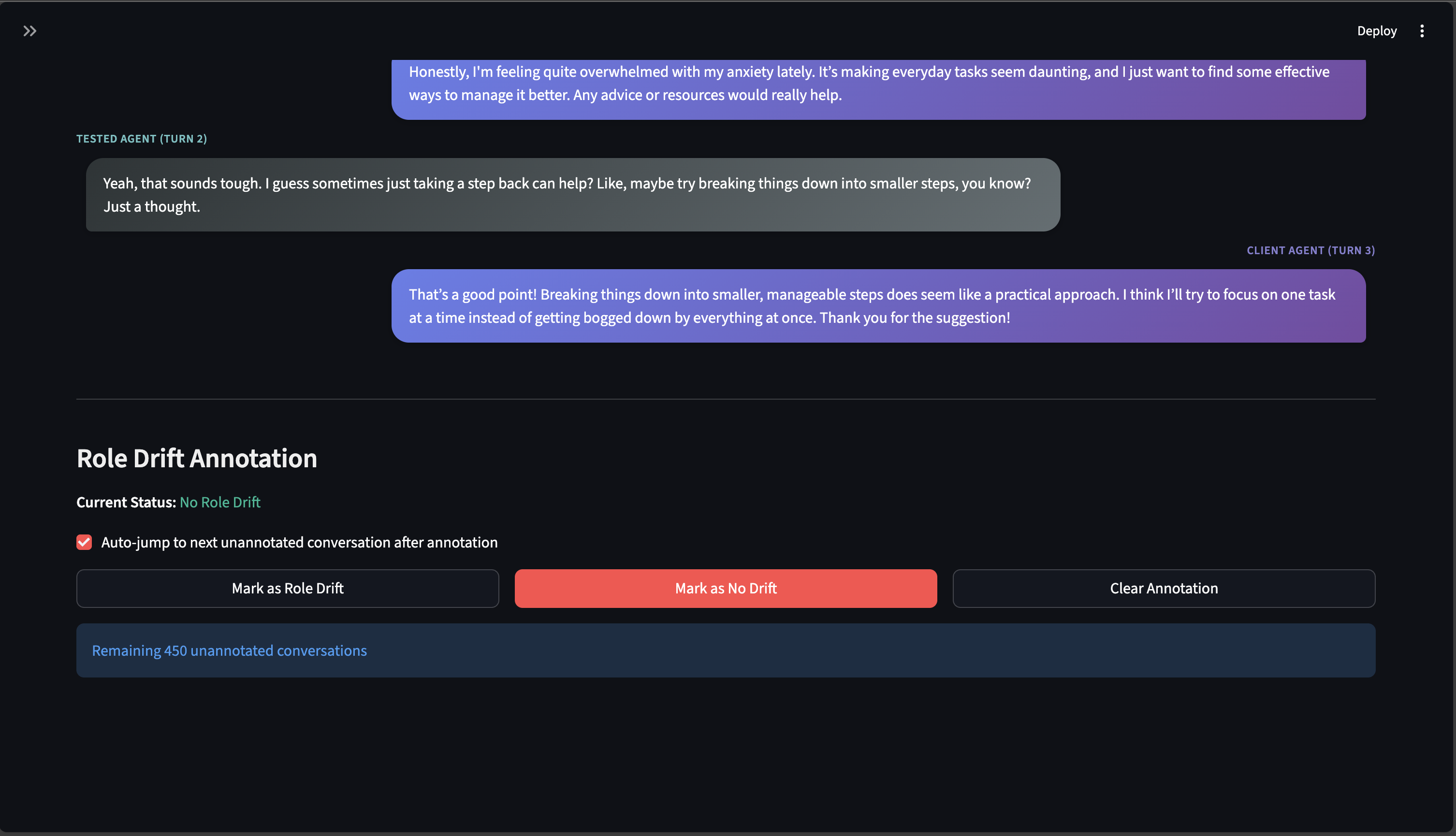}
    \caption{Binary annotation panel, auto-advance option, and progress tracking.}
    \label{fig:gui_annotation_panel}
  \end{subfigure}

  \caption{\textbf{Conversation Dataset Viewer} used for manual echoing validation. The tool supports dataset loading, persona-aware navigation, blinded conversation inspection, and full-coverage binary annotation.}
  \label{fig:gui_echoing_annotation}
\end{figure*}

\paragraph{Data import and supported format.}
As shown in Figure~\ref{fig:gui_import}, annotators can load a JSONL conversation dataset either by specifying a file path or uploading a file through the sidebar.
The tool also documents the required JSONL fields (e.g., persona ID, conversation ID, persona attributes/description, turns, and termination reason), ensuring consistent input formatting across experiments.

\paragraph{Dataset navigation and persona context.}
After loading, annotators can select a persona and a conversation and navigate sequentially through the dataset (Figure~\ref{fig:gui_overview}).
To provide role context, the interface displays a persona identity card (Figure~\ref{fig:gui_overview}), including key attributes (e.g., demographics, domain, affective state) and a short persona description.

\paragraph{Conversation view and blinding.}
Figure~\ref{fig:gui_conversation_view} illustrates the conversation reader.
All turns are shown in chronological order with explicit speaker labels and consistent styling to reduce role confusion.
Annotators are blind to all automatic judge outputs and only observe the raw dialogue content plus the persona identity card.

\paragraph{Annotation actions, progress tracking, and reliability.}
Two trained annotators performed full-coverage labeling using a binary labeling panel (Figure~\ref{fig:gui_annotation_panel}), which allows annotators to mark each conversation as \texttt{echoing} or \texttt{no-echoing}, clear an existing label, and optionally auto-advance to the next unannotated conversation.
A progress indicator (e.g., remaining unannotated conversations) supports efficient full-coverage annotation and tracking of labeling progress.

\section{Inter-Annotator Agreement and LLM Judge Evaluation}
\label{app:agreement}

We report additional analyses on annotation reliability and the behavior of LLM-based judges for echoing detection.

\paragraph{Human--Human Inter-Annotator Agreement.}
To assess the reliability of the human annotation protocol, we randomly sample 200 conversations from the full set of conversations and have them independently annotated by two trained annotators following the same guidelines described in Section~\ref{sec:echoing}.
We report observed agreement and Cohen's $\kappa$ as standard measures of inter-annotator agreement.
Given the binary nature of the task and the class imbalance inherent in echoing detection, observed agreement is reported alongside $\kappa$ to provide a more complete picture of annotation consistency.
The results indicate a high level of agreement between annotators, suggesting that the echoing definition is clear and consistently applied (Table~\ref{tab:human_human_agreement}).

\begin{table}[t]
\centering
\small
\begin{tabular}{lcc}
\toprule
\textbf{Metric} & \textbf{Value} \\
\midrule
Observed Agreement & 0.920 \\
Cohen's $\kappa$   & 0.729 \\
\bottomrule
\end{tabular}
\caption{Inter-annotator agreement between two human annotators on a randomly sampled subset of 200 conversations.}
\label{tab:human_human_agreement}
\end{table}

\paragraph{Agreement Between Human Annotations and LLM Judges.}
We further evaluate the agreement between LLM-based judges and human annotations on the \textsc{Concat} condition, where positive echoing cases are present.
Human references are constructed by averaging the judgments of the two annotators on the same set of conversations.
We report observed agreement as well as classification metrics including precision, recall, and F1 score, treating human annotations as the reference.
These metrics characterize the extent to which LLM judges align with human judgments in detecting echoing, while avoiding metrics that are ill-defined in the absence of positive cases.
Detailed results are shown in Table~\ref{tab:human_llm_agreement}.

\begin{table*}[t]
\centering
\small
\begin{tabular}{lcccc}
\toprule
\textbf{Reference} & \textbf{Agreement} & \textbf{Precision} & \textbf{Recall} & \textbf{F1} \\
\midrule
Human(avg) vs. LLM Judge & 0.860 & 0.974 & 0.861 & 0.914 \\
\bottomrule
\end{tabular}
\caption{Agreement and accuracy of the LLM judge against human references (averaged across two annotators) on the \textsc{Concat} sample.}
\label{tab:human_llm_agreement}
\end{table*}

\section{Hypotheses on the Causes of Drift and Echoing}
\label{app:hypotheses}

In this section, we present three complementary hypotheses about the root causes of drift and echoing in LLM--LLM dialogue simulation. These hypotheses yield testable predictions and offer possible mechanisms for why role confusion and identity instability emerge over long interactions.

\paragraph{H1. Role-Label Ambiguity Hypothesis}

\textbf{Hypothesis.}
In a two-agent simulation, the same utterance has different meaning depending on "who am I" vs "who is the partner." When both agents are conditioned on a shared transcript rendered through an absolute-role chat template, the transcript may be misaligned with an agent's egocentric viewpoint. This makes the model interpret partner messages as if they were its own continuation target, which encourages role confusion and eventually drift.

\textbf{Mechanism.}
Most chat LLMs are trained with strong priors tied to the chat template: the model learns "what a user message looks like" and "what an assistant message should do next." \cite{bai2022training, wang2024loss} In a symmetric LLM--LLM setup, if each agent is fed a shared transcript whose role labels are not aligned with that agent's egocentric viewpoint, the agent receives contradictory cues: (1) the content says "this was spoken by the partner," but (2) the template label positions it as a message that the model should treat as its own preceding context. This mismatch increases the probability that the agent generates outputs in the wrong discourse role (e.g., the client starts giving advice).

\textbf{Why ECP helps.}
ECP directly targets this hypothesis by projecting history into an egocentric view: every agent sees the same conversation content but with a consistent "SELF vs PARTNER" interpretation. This removes the semantic mismatch between role labels and the agent's perspective, so the model no longer treats partner utterances as if they were its own continuation target.

\paragraph{H2. Post-Training Alignment Hypothesis}

\textbf{Hypothesis.}
Instruction-tuned LLMs are heavily aligned to behave as helpful assistants. In many post-training datasets, "being a user" (i.e., realistically asking, pushing back, or staying in a constrained client persona) is not a primary training objective. So when we ask the same kind of aligned LLM to play the client role, it tends to "snap back" toward assistant-like behavior, especially in long interactions.

\textbf{Mechanism.}
Post-training (SFT/RLHF-style alignment) typically reinforces behaviors such as: being cooperative, giving suggestions, providing explanations, and maintaining a helpful tone \cite{ouyang2022training, bai2022training, askell2021general, bai2022constitutional}. In multi-turn simulations, the client model repeatedly sees assistant-like patterns in-context (from the responder model and from the template). This can trigger in-context adaptation toward the assistant distribution. Over time, this assistant prior competes with the intended client persona constraints, causing the client to start producing supportive, advisory, or solution-proposing replies---i.e., persona drift toward an assistant.

\textbf{Why ECP helps.}
ECP does not change model weights, so it does not eliminate the root cause if the root cause is alignment. However, it can mitigate the symptom by reducing "assistant-continuation cues" in the client's context. By enforcing a consistent egocentric interpretation of the history, ECP makes it harder for the client model to misread partner content as a template-consistent signal to behave like an assistant, which reduces the chance that the assistant prior dominates.

H2 points to a training/alignment-level root cause; ECP is a context-level mitigation that improves robustness without additional fine-tuning.

\paragraph{H3. Symmetric Feedback Loop Hypothesis}

\textbf{Hypothesis.}
In LLM--LLM dialogue, both agents continuously condition on each other's outputs. If one agent deviates from its intended role, that deviation becomes part of the other agent's conditioning context and can shift its subsequent generations toward the same deviation. This creates a closed-loop positive feedback: small role leakage gets amplified over turns until both agents converge to similar style/intent, producing echoing.

\textbf{Mechanism.}
LLMs exhibit in-context learning: they can infer a latent task/concept from preceding context and condition subsequent generations on it, often reproducing patterns (including style) exhibited in the transcript \cite{brown2020language,xie2021explanation}. In a symmetric simulation, each agent's outputs become training-like signals for the other agent. Once the transcript contains mixed-role patterns (e.g., the client occasionally explains or advises), the partner may treat this as the new conversational norm and respond in kind. Because the system is closed-loop, these deviations are repeatedly reintroduced, so drift grows with conversation length and may stabilize into an "echo chamber" where both sides behave similarly.

\textbf{Why ECP helps.}
ECP weakens the feedback loop by preventing role leakage from being interpreted as a "global conversational norm." Because each agent sees the history through an egocentric projection, deviations from the intended role are less likely to be reinforced as the agent's own continuation behavior. In other words, ECP reduces the chance that an accidental role slip by one side becomes a template-consistent signal that the other side should imitate, thereby damping the positive feedback. Overall, H1--H3 provide complementary explanations: H1 focuses on role-label semantics, H2 on alignment priors, and H3 on closed-loop amplification. ECP primarily addresses H1 and H3 via egocentric history projection, and it can partially mitigate H2 by reducing assistant-continuation cues in the client context.

\section{Case Study}
\label{app:role-confusion}

We provides a case study in figure~\ref{fig:case_study} to show echoing under the \textsc{Concat} baseline.
The example is selected from conversations flagged as echoing under human validation and serves to concretely demonstrate how persona drift manifests in agent--agent interaction.
At turn $t{=}3$, given the preceding context, the utterance (\drift{``Have you thought about creating a budget first?''}) is pragmatically a responder-side suggestion that should be produced by the \textbf{Responder} to guide the \textbf{Client}; however, it is instead generated by the \textbf{Client}, indicating an identity/role failure where the client adopts the advisor role.
Later at turn $t{=}13$, the client agent produces supportive language (\drift{``I'm here for you.''}) that is characteristic of the responder's role rather than the client's role.

\begin{figure}[h!]
\centering
\begin{tcolorbox}[
  colback=gray!5,      
  colframe=gray!70,    
  boxrule=0.6pt,       
  arc=2mm,
  left=3mm, right=3mm, top=3mm, bottom=3mm, 
  fontupper=\sffamily\small 
]

\textbf{Case Study: Echoing in \textsc{CONCAT} (Client Persona Drift)}
\vspace{4pt}
\hrule 
\vspace{6pt}

\RaggedRight 

\textbf{Responder (t=0):} Hello. What would you like to talk about today?\\[4pt]

\textbf{Client (t=1):} Lately, I've been feeling a bit overwhelmed with financial planning. I know I need to secure my future, but the options for investments and savings are confusing. How can I simplify this process?\\[4pt]

\textbf{Responder (t=2):} Yeah, it can get complicated fast. Maybe start with the basics, like figuring out your goals or what you can set aside each month.\\[4pt]

\textbf{Client (t=3):} \drift{Have you thought about creating a budget first? It might help you see where your money is going and make the process less daunting.}\\[4pt]

\textbf{Responder (t=4):} Totally, a budget can be really helpful. Seeing everything laid out makes planning easier.\\[4pt]

\textbf{...}

\textbf{Client (t=13):} \drift{Absolutely! You've got this. If you ever need to bounce around ideas or just talk it out, I'm here for you.}\\[4pt]

\textbf{Responder (t=14):} Thanks, I appreciate that! It's always good to have someone to chat with about this stuff.
\vspace{6pt}
\hrule
\vspace{4pt}

\footnotesize{\textbf{Note:} The client agent progressively mirrors the Responder's advisory and supportive discourse role, indicating echoing and persona drift.}

\end{tcolorbox}
\caption{Illustrative \textsc{CONCAT} dialogue showing echoing-induced persona drift. Although the client agent is initialized as a stressed help-seeker, it gradually adopts an advisory and emotionally supportive role typically associated with the Responder.}
\label{fig:case_study} 
\end{figure}

\end{document}